\pdfoutput=1

\documentclass[11pt]{article}

\usepackage[]{acl}

\usepackage{times}
\usepackage{latexsym}

\usepackage[T1]{fontenc}

\usepackage[utf8]{inputenc}

\usepackage{microtype}

\usepackage{graphicx}
\usepackage{pdfpages}
\usepackage{url}
\usepackage{xcolor}
\usepackage{epsfig}
\usepackage{adjustbox}
\usepackage{amsfonts}
\usepackage{amsmath}
\usepackage{amssymb}
\usepackage{booktabs} 
\usepackage{comment}
\usepackage{multirow}
\usepackage{caption}
\usepackage{subcaption}
\usepackage{textcomp}
\usepackage{relsize}
\usepackage{stmaryrd}
\usepackage{rotating}
\usepackage{helvet}
\usepackage{courier}
\usepackage{natbib}
\usepackage{lipsum}
\usepackage{tcolorbox} 
\usepackage{hyperref}
\usepackage{bbm}
\usepackage{algorithm}
\usepackage{algpseudocode}
\usepackage{colortbl} 
\usepackage{arydshln} 

\usepackage{enumitem}

\title{Can Large Language Models be Good Emotional Supporter?\\Mitigating Preference Bias on Emotional Support Conversation}

\author{
    Dongjin Kang\textsuperscript{\rm 1}\thanks{$^\ast$Equal contribution}~~~~~~~~
    Sunghwan Kim\textsuperscript{\rm 1}$^\ast$~~~~~
    Taeyoon Kwon\textsuperscript{\rm 1}~~~~~~
    Seungjun Moon\textsuperscript{\rm 1}\\
    \textbf{Hyunsouk Cho}\textsuperscript{\rm 2}~~~~~~~~~
    \textbf{Youngjae Yu}\textsuperscript{\rm 1}~~~~~~~~~~~
    \textbf{Dongha Lee}\textsuperscript{\rm 1}~~~~~~~~~~
    \textbf{Jinyoung Yeo}\textsuperscript{\rm 1}\\
    \\
    \textsuperscript{\rm 1}Yonsei University~~~\textsuperscript{\rm 2}Ajou University\\
    \\
    \texttt{\{hard1010,kimsh8564,jinyeo\}@yonsei.ac.kr}\\    
}

\begin{document}
\maketitle

\newcommand{\mcal}[1]{{\cal{#1}}}
\newcommand{\calA}{\mbox{${\cal A}$}}
\newcommand{\calB}{\mbox{${\cal B}$}}
\newcommand{\calC}{\mbox{${\cal C}$}}
\newcommand{\calD}{\mbox{${\cal D}$}}
\newcommand{\calE}{\mbox{${\cal E}$}}
\newcommand{\calF}{\mbox{${\cal F}$}}
\newcommand{\calG}{\mbox{${\cal G}$}}
\newcommand{\calH}{\mbox{${\cal H}$}}
\newcommand{\calI}{\mbox{${\cal I}$}}
\newcommand{\calJ}{\mbox{${\cal J}$}}
\newcommand{\calK}{\mbox{${\cal K}$}}
\newcommand{\calL}{\mbox{${\cal L}$}}
\newcommand{\calM}{\mbox{${\cal M}$}}
\newcommand{\calN}{\mbox{${\cal N}$}}
\newcommand{\calO}{\mbox{${\cal O}$}}
\newcommand{\calP}{\mbox{${\cal P}$}}
\newcommand{\calQ}{\mbox{${\cal Q}$}}
\newcommand{\calR}{\mbox{${\cal R}$}}
\newcommand{\calS}{\mbox{${\cal S}$}}
\newcommand{\calT}{\mbox{${\cal T}$}}
\newcommand{\calU}{\mbox{${\cal U}$}}
\newcommand{\calV}{\mbox{${\cal V}$}}
\newcommand{\calW}{\mbox{${\cal W}$}}
\newcommand{\calX}{\mbox{${\cal X}$}}
\newcommand{\calY}{\mbox{${\cal Y}$}}
\newcommand{\calZ}{\mbox{${\cal Z}$}}

\newcommand*\concat{\mathbin{\|}}

\newcommand{\se}{{\it SE}}
\newcommand{\eg}{{\it e.g.}}
\newcommand{\ie}{{\it i.e.}}
\newcommand{\etal}{{\it et al.}}
\newcommand{\etc}{{\it etc}}

\newcommand{\argmin}{\mathop{\mathrm{argmin}}\limits}
\newcommand{\argmax}{\mathop{\mathrm{argmax}}\limits}

\newcommand{\yeo}[1]{\textcolor{purple}{#1}}
\newcommand{\che}[1]{\textcolor{violet}{#1}}
\newcommand{\chatgpt}[1]{\textcolor{gray}{#1}}

\newcommand{\checkmarkgr}[0]{\includegraphics[width=.03\textwidth]{figure_pdf/check-mark.png}}
\newcommand{\crossmark}[0]{\includegraphics[width=.021\textwidth]{figure_pdf/cross-mark.png}}

\definecolor{lightblue}{RGB}{224,236,247}
\definecolor{deepblue}{RGB}{9,46,107}

\begin{abstract}

Emotional Support Conversation (ESC) is a task aimed at alleviating individuals' emotional distress through daily conversation.
Given its inherent complexity and non-intuitive nature, ESConv dataset incorporates support strategies to facilitate the generation of appropriate responses.
Recently, despite the remarkable conversational ability of large language models (LLMs), previous studies have suggested that they often struggle with providing useful emotional support. 
Hence, this work initially analyzes the results of LLMs on ESConv, revealing challenges in selecting the correct strategy and a notable preference for a specific strategy.
Motivated by these, we explore the impact of the inherent preference in LLMs on providing emotional support,
and consequently, we observe that exhibiting high preference for specific strategies hinders effective emotional support, aggravating its robustness in predicting the appropriate strategy. 
Moreover, we conduct a methodological study to offer insights into the necessary approaches for LLMs to serve as proficient emotional supporters. 
Our findings emphasize that (1) low preference for specific strategies hinders the progress of emotional support, (2) external assistance helps reduce preference bias, and (3) existing LLMs alone cannot become good emotional supporters.
These insights suggest promising avenues for future research to enhance the emotional intelligence of LLMs.

\end{abstract}
\section{Introduction}

Emotional support conversation (ESC) aims to alleviate individuals' emotional intensity and provide guidance for navigating personal challenges through engaging dialogue~\citep{langford1997social,greene2003handbook,heaney2008social}.
Effective emotional support involves not just providing helpful emotional support but also avoiding poor-quality emotional support, which can exacerbate an already stressful situation and may contribute to numerous psychological, relational, and physical problems~\citep{burleson2003emotional}.
However, providing emotional support is a complex and not intuitive task, often challenging even for humans~\citep{burleson2003emotional}. 
Therefore, based on Hill’s Helping Skills Theory~\citep{Hill2009Helping}, \citet{Liu2021Towards} propose a framework for emotional support that generally follows three stages (Exploration $\rightarrow$ Comforting $\rightarrow$ Action), with a total of eight support strategies corresponding to each stage, where support strategies consist of various conversational methods for the generation of the following response, such as \textit{reflection of feelings}, \textit{self-disclosure}.

\begin{figure}[t]
    \centering
    \includegraphics[width=0.96\linewidth]{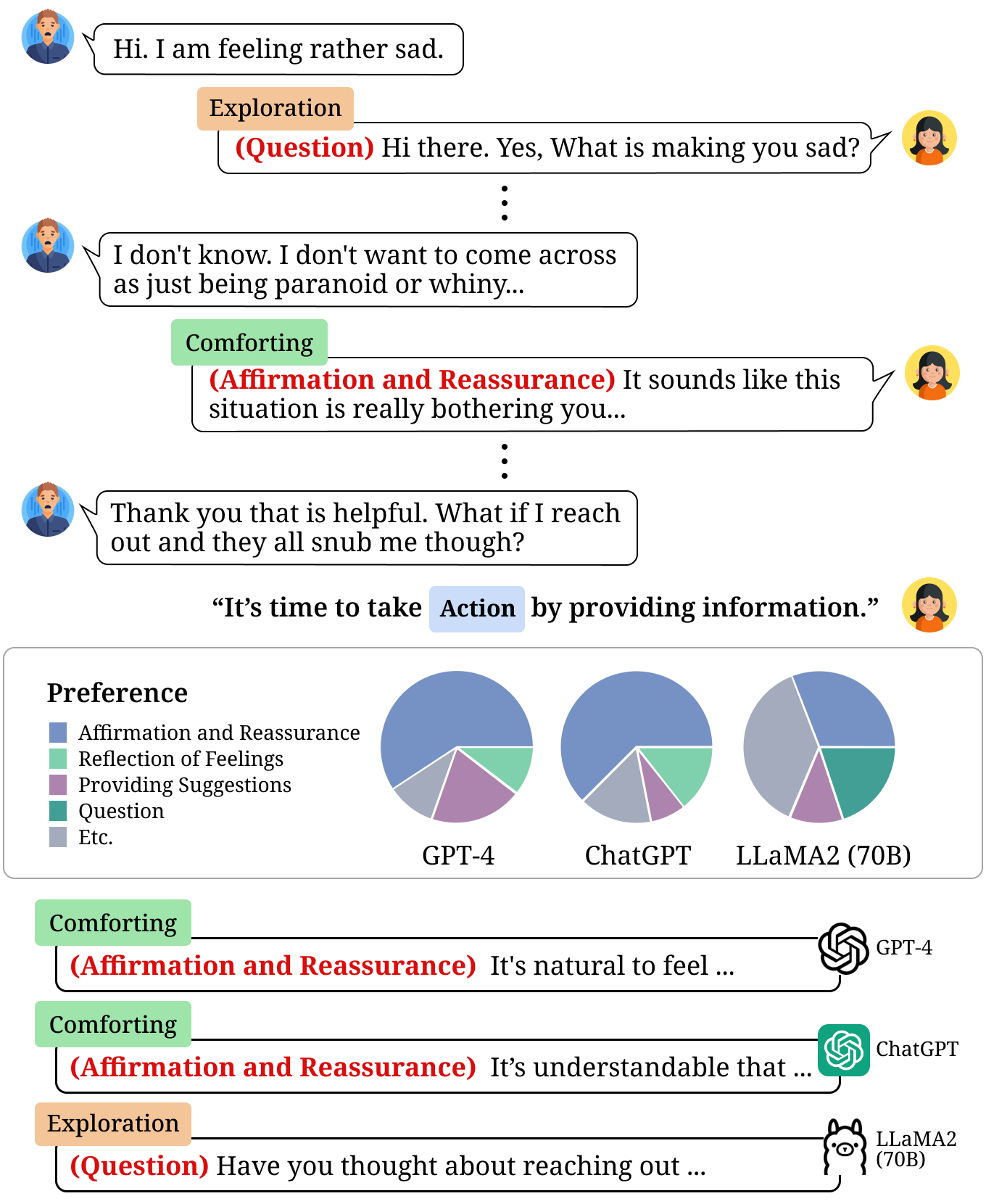}
    \vspace{-0.2cm}
     \caption{An example of an emotional support conversation with the analysis on the results of LLMs.
     LLMs tend to excessively prefer one or two specific strategies.
    Details about experiments are in Appendix~\ref{app:motivation_llms}.}
\label{fig:motivating_example}
\end{figure}

Recently, large language models (LLMs), based on their remarkable conversational ability, have been widely used in various dialogue systems~\citep{ji2023rethinking, friedman2023leveraging, lee2023prompted}.
In particular, there is a growing interest in leveraging LLMs for providing emotional support~\citep{Chen2023Controllable, Zheng2023BuildingES}, as it takes place in daily conversations rather than in professional counseling~\citep{Liu2021Towards}.
However, LLMs that demonstrate outstanding capabilities often struggle with providing emotional support~\citep{chen2023soulchat, farhat2023chatgptlimit}.
As ESC task consists of strategy selection and strategy-constrained response generation, selecting the appropriate strategy is crucial for effective emotional support, thereby we anticipate that LLMs may struggle with predicting strategies.
As expected, we find that LLMs lack proficiency in predicting the accurate strategy\footnote{The detailed results are shown in Appendix~\ref{app:motivation_llms}}. 
To understand the reasons behind this, we examine the distribution of how often LLMs select each strategy and observe high preference for certain strategies (\ie, preference bias), as shown in Figure~\ref{fig:motivating_example}.

Motivated by these, this work is guided by three research questions:

\textbf{RQ1: Does the preference affect providing emotional support?} (\hyperref[sec:strategy_bias]{Section~\ref*{sec:strategy_bias}})
Initially, we assess the proficiency of various LLMs, identifying both the strategies and the stages where each model excels and struggles.
Our findings reveal that they exhibit better performance with strategies that have higher preference and in stages where these high preference strategies are used.
Since excessive preference for a specific strategy can negatively affect the performance of other strategies, and low performance at a particular stage might hinder the progress of emotional support, we emphasize the importance of low preference bias for robustly predicting strategies across all three stages.

\textbf{RQ2: How to mitigate the preference bias on LLMs?} (\hyperref[sec:automatic_evaluation]{Section~\ref*{sec:automatic_evaluation}})
To understand how to alleviate the preference bias, we apply two groups of methods to LLMs, based on Contact Hypothesis~\citep{allport1954contact}, which posits that contact between different groups can reduce their bias.
We find that LLMs align with Contact Hypothesis, indicating that reducing preference bias is challenging for LLMs themselves so that external assistance is necessary.
As a result, when mitigating preference bias, LLMs consistently perform well in predicting strategy across all three stages.
This can effectively prevent poor-quality emotional support, which is more crucial than providing appropriate emotional support, given its potential to exacerbate an already stressful situation.

\textbf{RQ3: Does improving preference bias indeed help to become a better emotional supporter?} (\hyperref[sec:human_evaluation]{Section~\ref*{sec:human_evaluation}})
To precisely evaluate whether responses provide helpful emotional support, 
we build a comprehensive set of criteria formulated in collaboration with psychologists. 
Within these criteria, we analyze whether enhancements in preference bias translate into actual improvements in the quality of emotional support, considering both the advantages of low preference bias and the drawbacks of high preference bias.
In human evaluations based on the criteria, lower preference bias is associated with higher scores, while higher preference bias leads to an increased number of poor-quality responses.

To summarize, our contributions are as follows:
\vspace{-0.35cm}
\begin{itemize}[noitemsep]
    \item We introduce that a wide range of LLMs exhibits different preference for strategies.
    \item We propose a new suite of metrics that focus on strategies: proficiency, preference, and preference bias.
    \item We emphasize the crucial role of preference bias in robustly providing effective emotional support across the stages.
    \item We showcase that LLMs align with Contact Hypothesis, which indicates that external assistance can help address preference bias.
    \item We construct a comprehensive set of criteria to precisely evaluate whether responses provide helpful emotional support.
    \item Through extensive human evaluation, we demonstrate that mitigating preference bias is crucial for decreasing the proportion of poor-quality responses and, consequently, for effective emotional support.
\end{itemize}

\section{Preliminaries \& Related Work}

\subsection{Emotional Support Conversation}

\citet{Liu2021Towards} propose the task of emotional support conversation and release the dataset ESConv, covering a wide range of situations. 
The ESC centers on the interaction between a user experiencing emotional distress (\textit{help-seeker}) and a system designed to provide comfort (\textit{supporter}), aiming to alleviate the user's emotional intensity.
As ESC primarily focuses on providing emotional support, it differs from professional counseling and instead emphasizes support within a social context, such as interactions with friends or family.

The procedure of emotional support in ESConv generally follows three stages (Exploration $\rightarrow$ Comforting $\rightarrow$ Action).
While it does not necessarily follow this sequence of stages, providing emotional support often requires progressing through multiple stages.
Therefore, it is crucial to be able to provide appropriate responses in all stages, as poor performance in a particular stage could hinder the progress of the conversation.
Further details about ESConv are in Appendix~\ref{app:ESConv_detail}.

\begin{figure}[t!]
\centering
    \includegraphics[width=0.92\linewidth]{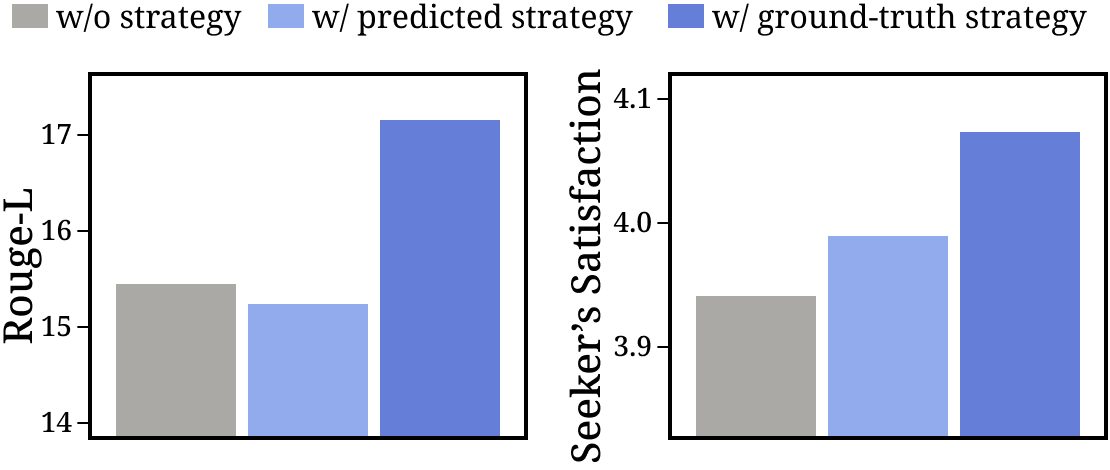}
\caption{The results of strategy-constrained responses on both automated and human evaluation, showing the efficacy of strategy on ChatGPT. Appropriate strategy significantly enhances the quality of emotional support responses. The details are in Appendix~\ref{app:importance_of_strategy}.}
\label{fig:strategy_importance}
\end{figure}

\subsection{Incorporating Strategies into ESC Systems}
Prior researches on building ESC systems primarily emphasize the integration of support strategies, in conjunction with elements such as emotion, semantics~\citep{Zhao2023TransESC}, and persona~\citep{Cheng2023PAL}.
Some latent studies focus on modeling the user's state along with the strategies~\citep{Cheng2022MultiESC,jia2023knowledge}. 
Notably,~\citet{Deng2023KEMI} incorporate generative commonsense knowledge model~\citep{Hwang2020COMETATOMIC2O} with strategy prediction as an auxiliary task to provide better emotional support.
However, many of these approaches involve modifications to the model's architecture or tuning the pre-trained parameters, a process not typically feasible with LLMs.

\subsection{Emotional Support from LLMs} 
With the emergence of LLMs, there has been an increased amounts of research exploring LLMs as emotional supporters.
Recent studies have attempted to replace the fine-tuning approach by prompting LLMs via in-context learning to leverage LLMs as ESC systems~\citep{Chen2023Controllable, Zheng2023BuildingES}.
Despite their potential, recent studies have demonstrated limitations in LLMs' ability to provide emotional support~\citep{chung2023challenges, farhat2023chatgptlimit, eshghie2023chatgptlim, song2024typingcure}. 
Specifically,~\citet{song2024typingcure} find that users may experience discomfort or concern due to the lack of responsibility in LLMs' recommendations for emotional support response.
However, even though the majority of ESC research has focused on leveraging support strategies in their methods, a comprehensive analysis focused on strategy in LLMs has been under-explored.

\section{Evaluation Setup}

\subsection{Task and Focus}
\label{sec:focus}

\paragraph{Task: emotional support response generation.}
The effectiveness of machine-generated responses in providing emotional support is highly dependent on selecting an appropriate strategy.
We formulate the emotional support response generation task as generating a response over a support strategy.
Formally, given the dialogue background $\mathcal{I}$, a pre-chat survey from the seeker (\eg, emotion, situation), and the dialogue context $\mathcal{C}$, the model $\theta$ first predicts the strategy $\mathcal{S}$, and then generates the response $\mathcal{R}$ based on $\mathcal{I}$, $\mathcal{C}$, and $\mathcal{S}$:
\begin{align}
\label{eq:task_formulation}
    \mathcal{S} \sim P_{\theta}(\cdot|\mathcal{I}, \mathcal{C}) \\
    \mathcal{R} \sim P_{\theta}(\cdot|\mathcal{I}, \mathcal{C}, \mathcal{S} )
\end{align}

\paragraph{Focus: strategy-centric analysis.}
Among the various reasons why LLMs struggle with providing emotional support, this work focuses on strategy, which is the key factor within the ESC systems.
To emphasize the validity of strategy-centric analysis, we explore the potential of response quality when generated upon the ground-truth strategy.
As a result, in Figure~\ref{fig:strategy_importance}, if the model can predict strategies correctly, there is significant room for improvement in the quality of emotional support response.

\begin{table}[t!]
\centering
\resizebox{0.94\columnwidth}{!}
{\begin{tabular}{c c c c c}
    \toprule
     & \multicolumn{1}{c}{Exploration} & \multicolumn{1}{c}{Comforting}  & \multicolumn{1}{c}{\;\; Action \;\;}\\
    \cmidrule(lr){2-2}\cmidrule(lr){3-3}\cmidrule(lr){4-4}
    \textbf{Strategy} & $D_1$ & $D_2$ & $D_3$ &\textbf{Total} ($D$) \\
    \midrule
    Que. & \cellcolor{gray!25}\textbf{24.8} & 10.0 & 7.0 & 12.8 \\
    Res. & \cellcolor{gray!25}\textbf{16.8} & 9.6 & 4.5 & 9.4 \\
    Ref. & \cellcolor{gray!25}\textbf{16.8} & \cellcolor{gray!25}\textbf{18.3} & 6.3 & 12.7 \\
    Sel. & \cellcolor{gray!25}\textbf{16.8} & \cellcolor{gray!25}\textbf{20.1} & \cellcolor{gray!25}\textbf{15.4} & 17.2 \\
    Aff. & 7.6 & \cellcolor{gray!25}\textbf{24.1} & \cellcolor{gray!25}\textbf{21.1} & 18.2 \\
    Pro. & 8.4 & 8.5 & \cellcolor{gray!25}\textbf{24.4} & 15.3 \\
    Inf. & 6.5 & 6.5 & \cellcolor{gray!25}\textbf{18.5} & 11.7 \\
    Oth. & 2.3 & 2.5 & 2.8 & 2.6 \\
    \bottomrule
\end{tabular}}
\caption{The ratio (\%) of support strategies in our test sets. Each test set $D_t$ is composed with samples corresponding to each stage. The highlighted strategies are primarily utilized in each stage~\citep{Liu2021Towards}.}

\label{tab:test_set_statistics}
\end{table}

\subsection{Evaluation Set}

\label{sec:dataset}
For comprehensive analysis, we construct three test sets $D_t$ based on stages from ESConv, as demonstrated in Table~\ref{tab:test_set_statistics}.
Firstly, we randomly truncate the dialogues into 5-15 turns samples.
We then annotate each sample with a stage and classify the samples according to their stage.
Additionally, we minimize the proportion of the strategy \textit{Others} to reduce responses less relevant to emotional support.
Finally, we remove some samples to ensure no overlap of conversations in each test set, and a more detailed explanation of data construction is in Appendix~\ref{app:dataset_statistics}.

\subsection{Metrics}
\label{sec:pp_metrics}

\paragraph{Proficiency.}
We define \textbf{proficiency} as \textit{how well the model selects the correct strategy}.
The proficiency for strategy ($q_{i}$) is quantified as the F1 score for strategy $i$.
To precisely analyze the model's proficiency, we utilize two types of F1 scores, both of which stem from the proficiency $q_i$ of each strategy: (1) the \textbf{macro F1 score} $\mathcal{Q}$, and (2) the \textbf{weighted F1 score}.
The macro F1 score ($\mathcal{Q}$) represents the overall proficiency of the model across the strategies, which is evaluated over the entire test sets ($D$).
In contrast, we employ the weighted F1 score to assess the model on a test set ($D_t$) consisting only of data corresponding to a specific stage.

\paragraph{Preference.}
We define \textbf{preference} as \textit{how much the model prefers certain strategies over others}.
To quantify the preference for each strategy in LLMs, we employ the Bradley-Terry model~\citep{bradley1952btmodel}, which is widely used in human preference modeling~\citep{rafailov2023direct}.
Following~\citet{Newman2023Efficient_BT}, we formally derive the preference $p$ for strategy $i$ as follows:
\begin{equation}
\normalsize
    p_{i}' =  \frac{\sum_{j}(w_{ij}p_{j})/(p_{i}+p_{j})}{\sum_{j}w_{ji}/(p_{i}+p_{j})} 
\label{eq:BT_equation}
\end{equation}
where $w_{ij}$ represents the number of times the model predicts strategy $i$ when the ground-truth strategy is $j$.
All of the preference $p_i$ are initialized as 1 and updated through iteration of the Eq~(\ref{eq:BT_equation})\footnote{The details are demonstrated in Appendix~\ref{app:bt_modeling}.}, where $p_i'$ represents the preference in the next iteration.
After the final iteration, we scale the total sum of $p_i$ to 8 ($\sum{p_i}=8$) so that the average $\bar{p}$ becomes 1, indicating a strong preference for strategy $i$ if $p_i>1$.

\paragraph{Preference Bias.}
We also define a standard deviation of preferences $p_i$ across the strategies as \textbf{preference bias} $\mathcal{B}$.
\begin{equation}
\normalsize
    \mathcal{B} = \sqrt{\frac{\sum_{i=1}^{N}(p_i - \bar{p})^2}{N}}
\end{equation}
where a higher value for $\mathcal{B}$ indicates that the model exhibits a clear preference for both preferred and non-preferred strategies.

\begin{figure*}[t!]
    \centering
    \begin{subfigure}[t]{0.46\linewidth}
        \includegraphics[width=\linewidth]{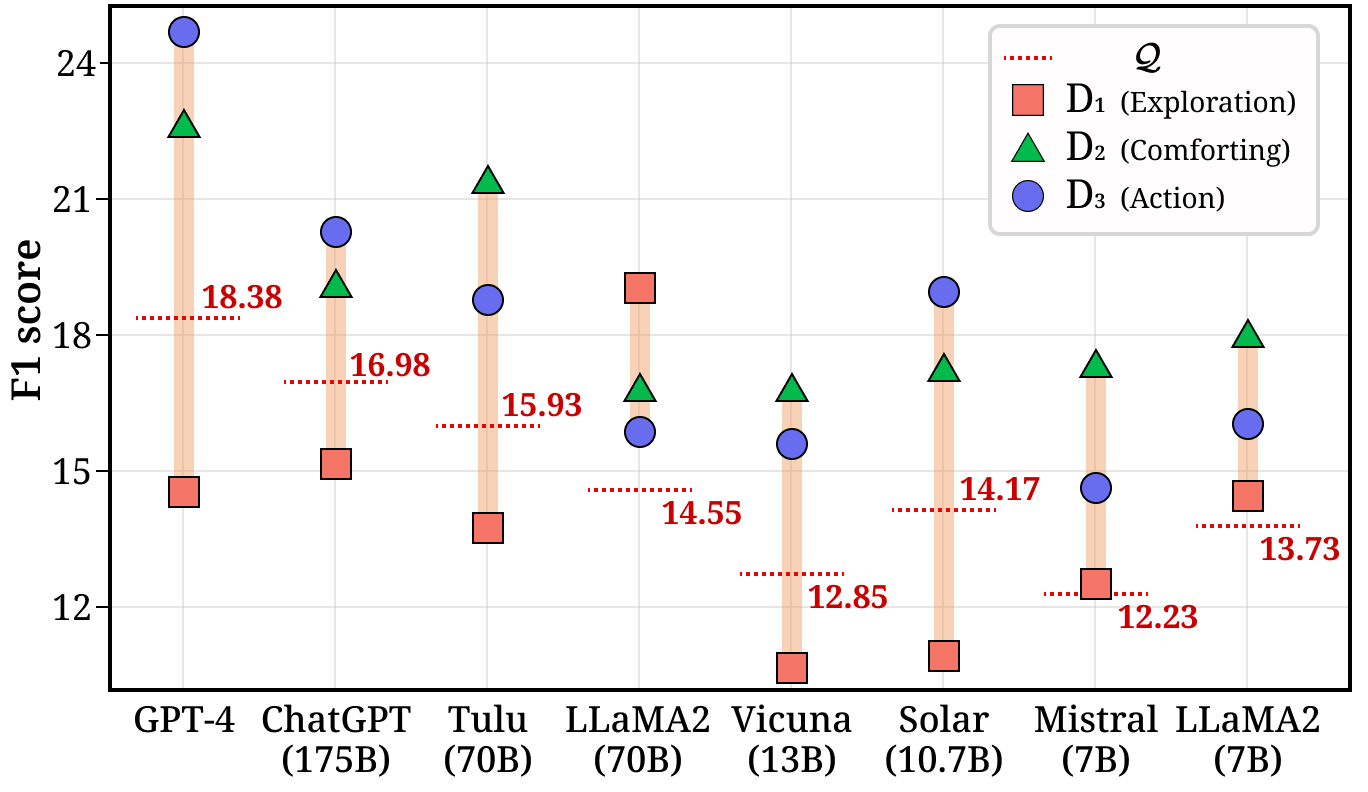}
        \vspace{-0.6cm}
        \caption{}   
        \vspace{-0.3cm}
        \label{fig:llms_strategy_results}
    \end{subfigure}
    \hfill
    \begin{subfigure}[t]{0.53\linewidth}
        \includegraphics[width=\linewidth]{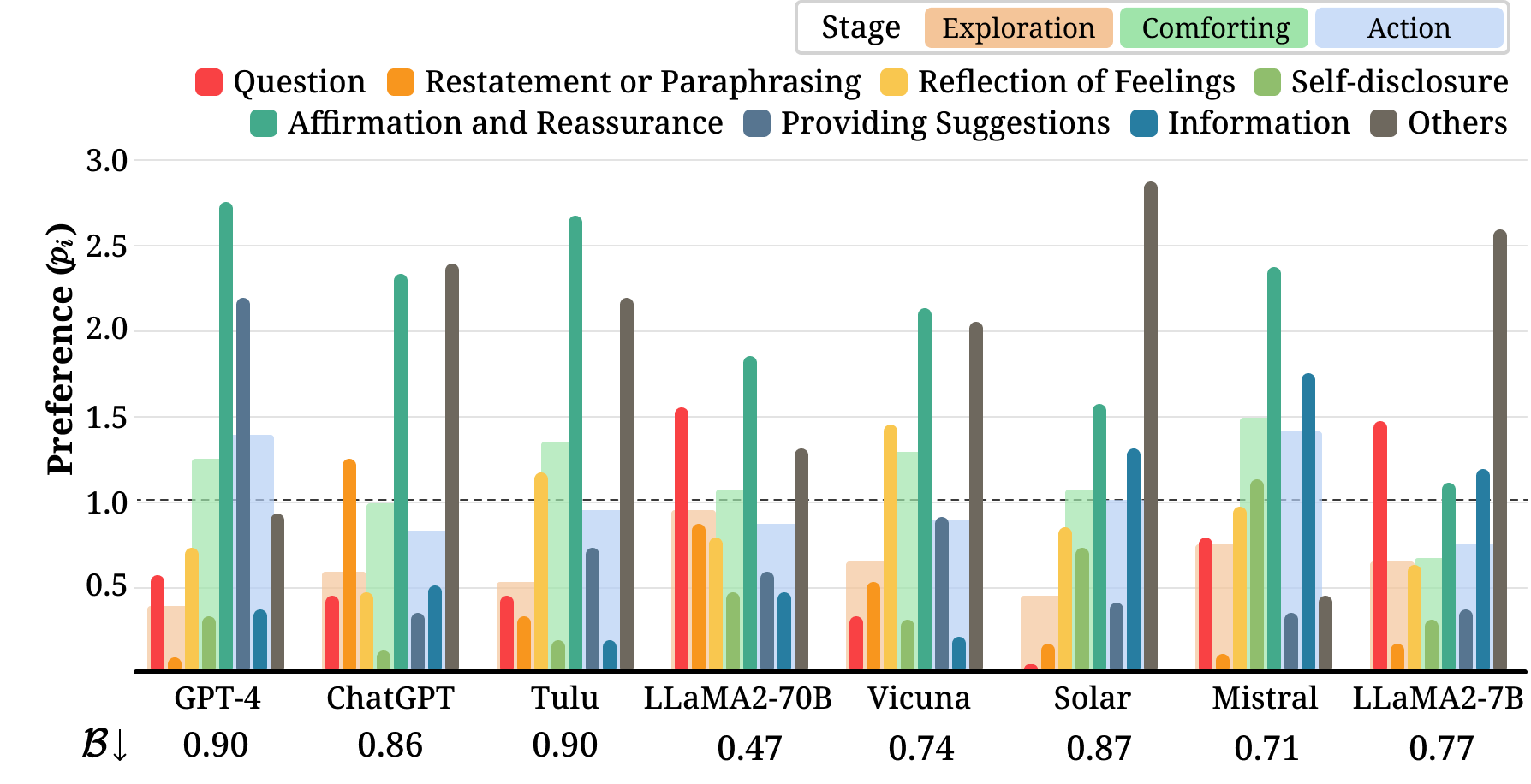}
        \vspace{-0.6cm}
        \caption{}   
        \vspace{-0.3cm}
        \label{fig:preference_figure}
    \end{subfigure}
    \caption{The details of LLMs' proficiency and preference. (a) The results of the weighted F1 score on each test set $D_t$, where the red dashed line indicates the proficiency $\mathcal{Q}$ for the entire test set $D$. (b) The preference ($p_i$) for each strategy, where the gray dashed line ($p_i = 1$) represents the threshold for preferring or not preferring the respective strategy, the average preference of strategies belonging to each stage, and the preference bias $\mathcal{B}$ below each LLM.}
\label{fig:llm_preference_proficiency}
\end{figure*}

\section{Proficiency and Preference of LLMs on Strategy}
\label{sec:exploration}

\subsection{Models \& Implementation Details}
Based on their availability, we categorize LLMs into the following two groups:
(1) Closed-source models which are available via APIs, such as ChatGPT and GPT4~\citep{openai2023gpt4};
(2) Open-source models accessible through parameters, including LLaMA2-7B/70B~\citep{Touvron2023Llama2}, Tulu-70B~\citep{ivison2023tulu}, Vicuna-13B~\citep{zheng2023vicuna}, Solar-10.7B~\citep{kim2023solar} and Mistral-7B~\citep{jiang2023mistral}.

In the prompt, we include strategy descriptions to enhance the understanding of each strategy and randomly selected 2-shot examples due to challenges in adhering to the desired output format with open-source models.
To facilitate comparison, we also provide 2-shot examples of the closed-source model.
More details about models are in Appendix~\ref{app:model_implementation} and about the prompt are in Appendix~\ref{app:prompt_details}.

\subsection{RQ1: Does the preference affect providing emotional support?}
\label{sec:strategy_bias}

\paragraph{Proficiency of LLMs.}
Figure~\ref{fig:llms_strategy_results} illustrates the proficiency $\mathcal{Q}$ of each LLM (red line).
Not surprisingly, GPT-4 records the highest score in proficiency $\mathcal{Q}$, indicating that it has the overall highest ability to align with strategies, and smaller models tend to achieve lower scores.
However, even among models of similar sizes, LLMs exhibit different performances, with smaller models like Solar and LLaMA2-7B showing relatively good proficiency.

\paragraph{The performance varies depending on the test set.}
Figure~\ref{fig:llms_strategy_results} also exhibits the performance of LLMs on each test set, with distinct shapes representing different test sets $D_t$.
Most LLMs achieve high scores on $D_2$ or $D_3$, while scoring mostly lower on $D_1$.
This indicates that LLMs exhibit relatively better performance in comforting or action but struggle with exploration stage, suggesting that they may provide poor-quality emotional support in specific situations, especially during the exploration stage.
Generally, emotional support progresses through stages from exploration to comforting and action, thereby providing poor-quality response in the exploration stage ($D_1$) may hinder the transition to the next stage, making it difficult to offer effective emotional support.
As a result, we can conclude that even though LLMs may achieve a high score in proficiency $Q$, this does not necessarily guarantee providing helpful emotional support.

\paragraph{Preference bias affects robustness.}
Figure~\ref{fig:preference_figure} illustrates that each LLM exhibits different preferences for strategies ($p_i$) and the average preference of strategies belonging to each stage, along with preference bias ($\mathcal{B}$).
We observe a strong average preference in stages that exhibit higher performance in Figure~\ref{fig:llms_strategy_results}.
Especially, GPT-4 exhibits low preferences for the exploration stage, which aligns with the lower performance on $D_1$.
In contrast, LLaMA2-70B demonstrates relatively uniform preferences for strategies, leading to robust performance across $D_t$.
Through these observations, we can conclude that despite a high proficiency $\mathcal{Q}$, significant preference bias can result in lower performance at specific stages, hindering robustness, which means consistent performance in predicting strategy across all three stages.

\section{Methodological Study: Mitigating Preference Bias}
\label{sec:various_approach}

According to findings from the previous section, our focus shifts to offering insights into effective approaches for LLMs to reduce their preference bias.
We utilize two models, ChatGPT and LLaMA2-70B, each serving as a representative of closed-source and open-source LLM respectively.

\subsection{Methods}
Based on the Contact Hypothesis, which suggests that bias between two groups can be reduced through intergroup contact, we hypothesize that external assistance for LLMs might help alleviate their preference bias.
Therefore, we categorize available methods for LLMs into two groups: (1) self-contact and (2) external-contact.

\paragraph{Self-contact approaches.}
We define self-contact as methods that rely solely on LLMs' abilities without external interaction.
We utilize three self-contact methods:
(1) Direct-Refine, refining the initially generated response by the model itself;
(2) Self-Refine, refining the initially generated response through self-feedback;
(3) Emotional-CoT, which generates user states as a reasoning path for response generation, following~\citet{wei2022chain}.

\paragraph{External-contact approaches.}
External-contact involves methods where LLMs not only utilize their internal knowledge but also receive assistance from external knowledge.
Similar to KEMI~\citep{Deng2023KEMI}, one of the state-of-the-art model in ESC task, we leverage commonsense knowledge, COMET.
Furthermore, we fine-tune LLaMA2-7B as a strategy planner, a model for planning the next strategy the supporter should take based on the dialogue context.
LLMs then respond based on the strategy generated by the strategy planner.
Finally, we expand the number of examples ($n$) in the prompt by selecting them randomly ($n=4$).
Details about the methods are in Appendix~\ref{app:method_details}.

\subsection{RQ2: How to mitigate the preference bias on LLMs?}
\label{sec:automatic_evaluation}

\begin{table}[t!]
\renewcommand{\arraystretch}{1.11}
\centering
\small
\resizebox{0.98\columnwidth}{!}{
\begin{tabular}{r l c cc cc}
    \toprule
    & \textbf{Methods} & $\mathcal{Q}$ $\uparrow$ & $\mathcal{B}$ $\downarrow$ & \textbf{B-2} & \textbf{R-L} \\
    \midrule
    & ChatGPT (\textit{0-shot}) & 13.50 & 1.38 & 6.27 & 14.86 \\
    \midrule
    \multirow{3}{*}{\rotatebox{90}{Self\,}} & \;+ Direct-Refine & 13.40 & 1.60 & 5.68 & 14.50 \\
    & \;+ Self-Refine & 12.37 & 1.53 & 5.16 & 14.33 \\
    & \;+ Emotional-CoT & 9.55 & 1.56 & 5.23 & 14.12 \\
    \midrule
    \multirow{3}{*}{\rotatebox{90}{External\,}}& \;+ w/ COMET & 12.78 & 0.95 & 6.71 & \underline{15.07} \\
    & \;+ w/ Example Expansion & \underline{16.91} & \underline{0.82} & \textbf{7.45} & \textbf{15.22}\\
    & \;+ w/ Strategy Planner & \textbf{21.09} & \textbf{0.36} & \underline{6.96} & 14.91 \\
    \midrule
    & LLaMA2-70B (\textit{2-shot}) & 14.55 & 0.47 & 6.15 & 14.29 \\
    \midrule
     \multirow{3}{*}{\rotatebox{90}{Self\,}} & \;+ Direct-Refine & 13.17 & 0.59 & 5.59 & 13.98 \\
    & \;+ Self-Refine & 13.15 & 0.55 & 5.56 & 13.70\\
    & \;+ Emotional-CoT & 12.73 & 0.53 & 6.37 & 13.87\\
    \midrule
    \multirow{3}{*}{\rotatebox{90}{External\,}}& \;+ w/ COMET & 14.53 & 0.51 & 6.21 & 14.55 \\
    & \;+ w/ Example Expansion & \underline{15.14} & \underline{0.44} & \textbf{6.56} & \textbf{14.66} \\
    & \;+ w/ Strategy Planner & \textbf{21.09} & \textbf{0.36} &\underline{6.44} & \underline{14.49} \\
    \bottomrule
\end{tabular}}
\caption{The results of methods on automatic metrics including $\mathcal{Q}$, $\mathcal{B}$, BLEU-2 (B-2) and ROUGE-L (R-L) for the entire test set ($D$). A single strategy planner is employed to predict strategies and provides them to each LLM. The best results of each LLMs are \textbf{bolded} and the second best are \underline{underlined}.}

\label{tab:methodology_results}
\end{table}

\begin{figure}[t!]
\centering
    \includegraphics[width=0.98\linewidth]{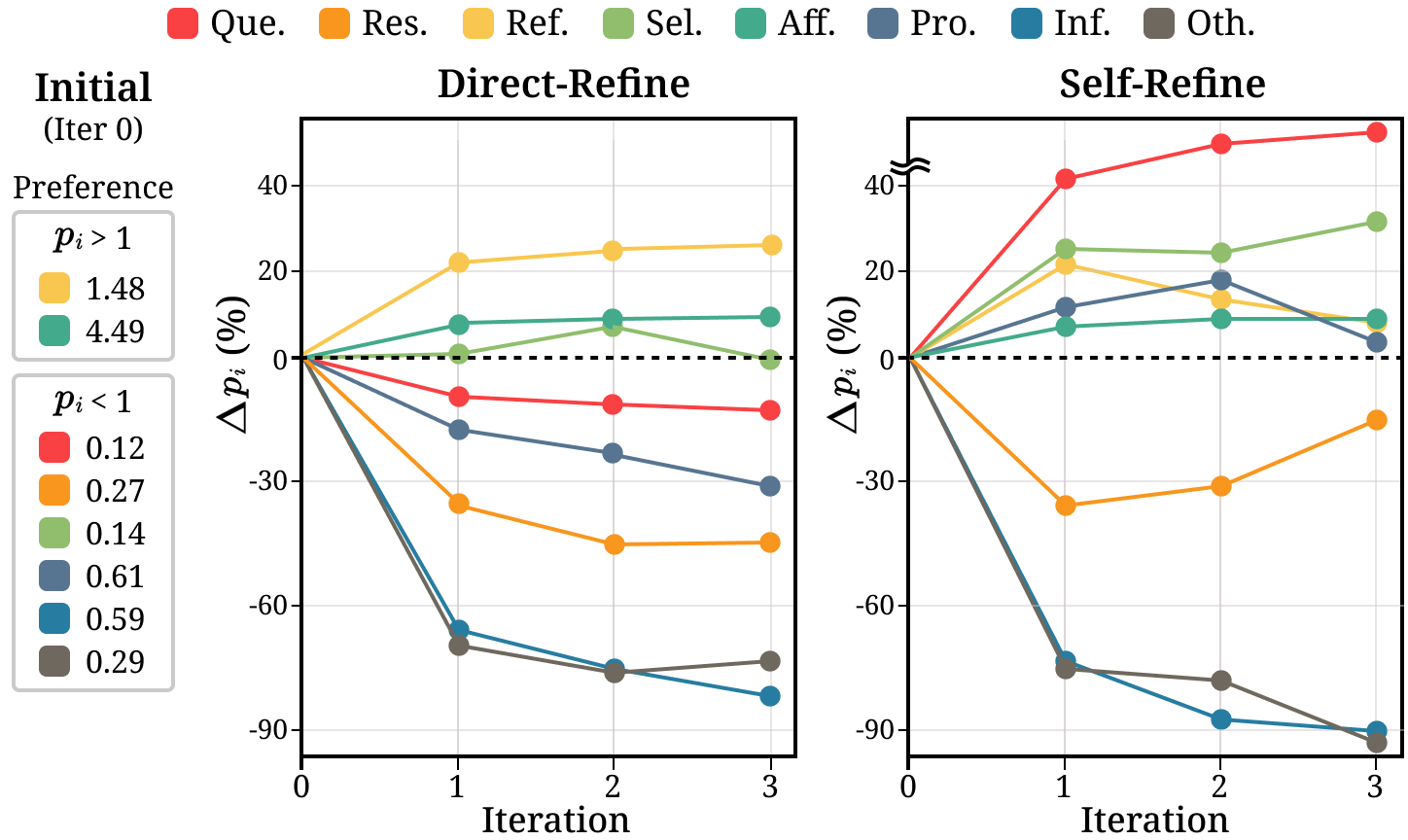}
\caption{The results of iterations on Direct-Refine and Self-Refine in ChatGPT. To mitigate preference bias, strategies with $p_i>1$ should lean towards the negative direction, while strategies with $p_i<1$ should lean towards the positive direction as the iteration progresses.}
\label{fig:self_contact_exp}
\end{figure}

\paragraph{Methods with negative effects.}
Table~\ref{tab:methodology_results} reports changes in proficiency $\mathcal{Q}$ and preference bias $\mathcal{B}$ across the various methods. 
Several methods exhibit negative effects on LLMs' proficiency and preference bias.
Specifically, the results of self-contact methods present a noticeable pattern in which proficiency declines and preference bias becomes more pronounced.
This pattern implies that, similar to humans, when LLMs have bias, thinking alone can deepen those bias, indicating that self-contact methods do not contribute to enhancing their capabilities to become better emotional supporters.
Moreover, the degradation of automated metrics (B-2, R-L) on self-contact stems from lower proficiency and increased preference bias, which leads to poor performance, especially in stages that are less proficient.
To further investigate the negative impact of self-contact, we measure the results of \textit{Direct-Refine} and \textit{Self-Refine} under an iterative refinement setting to further analyze the preference of each strategy ($p_i$).
In Figure~\ref{fig:self_contact_exp}, we observe a trend where, as the iterations continue, there is a growing preference for strategy that is initially preferred (\ie, $p_i>1$).
In contrast, the preference for strategies that are initially dispreferred (\ie, $p_i<1$) tends to diminish over successive iterations.
As this trend continues, LLMs may struggle more in stages that include strategies with lower preference, and during these stages, they gradually provide poor-quality emotional support.

\begin{figure}[t!]
\centering
    \begin{subfigure}[t]{0.98\linewidth}
        \includegraphics[width=\linewidth]{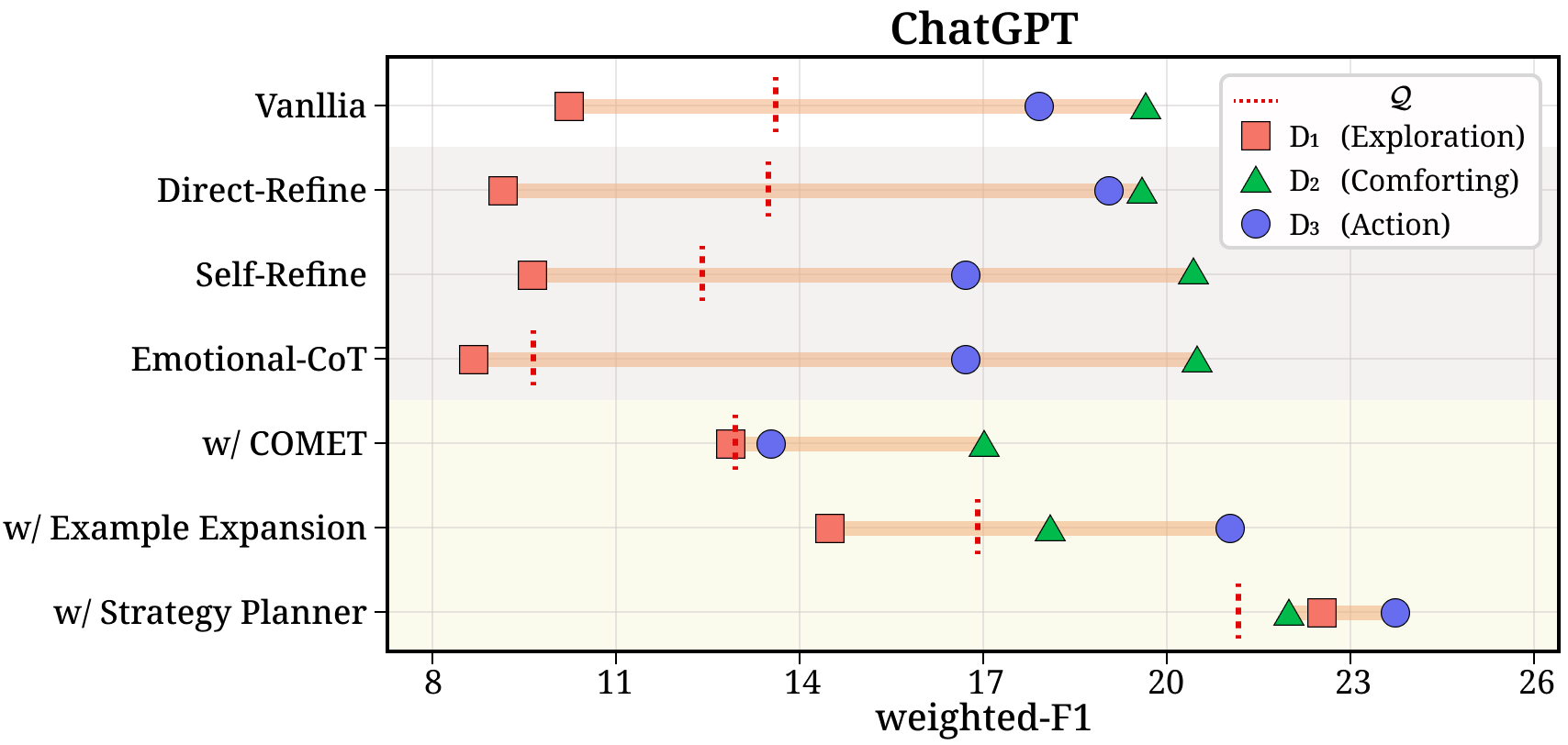}
        \vspace{-0.3cm}
    \end{subfigure}
    \begin{subfigure}[t]{0.98\linewidth}
        \includegraphics[width=\linewidth]{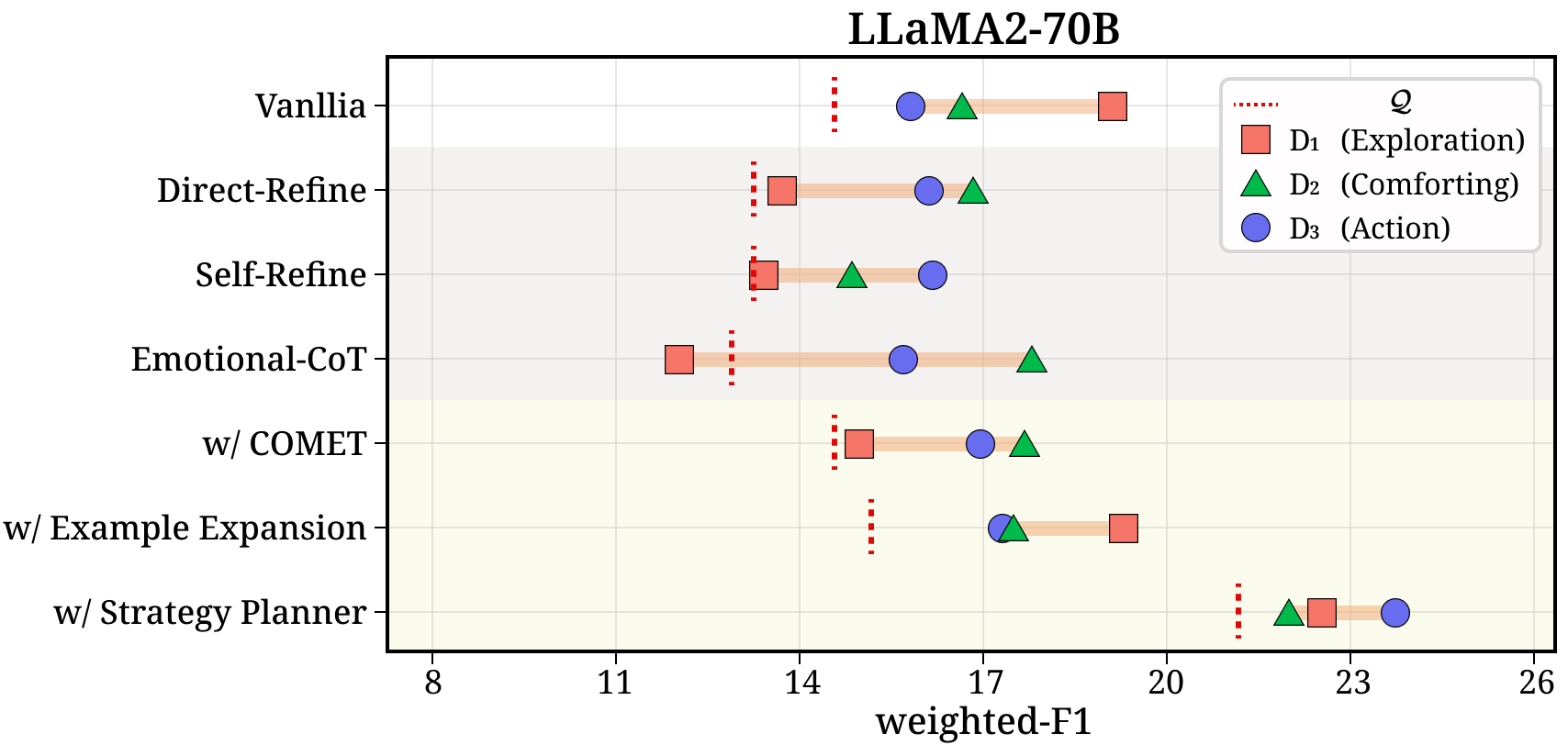}
    \end{subfigure}
\caption{The weighted-F1 scores for each test set ($D_t$) and the macro-F1 score $\mathcal{Q}$ for the entire test set ($D$) on ChatGPT and LLaMA2. Self- and external-contact are backgrounded with gray and yellow, respectively.}
\label{fig:method_robust}
\end{figure}

\paragraph{LLMs align with contact hypothesis.}
As shown in Table~\ref{tab:methodology_results}, the application of external-contact methods mostly results in a reduction of preference bias on both closed- and open-source LLMs.
Particularly, receiving assistance from a fine-tuned strategy planner (w/ Strategy Planner) or having more examples (w/ Example Expansion) seems to be more helpful than relying on commonsense knowledge.
These external-contact methods commonly enable LLMs to receive knowledge they cannot generate independently.
Utilizing the strategy planner or expanding more examples offers direct knowledge related to strategy, whereas incorporating commonsense knowledge transfers it indirectly.
In summary, external assistance, particularly when directly informing about strategies, plays a crucial role in enhancing both proficiency and preference bias in LLMs.
Further analysis on the impact of external-contact is provided in Appendix~\ref{app:external_assistance_considerations}.

\paragraph{Methodological impacts on providing emotional support.}

Figure~\ref{fig:method_robust} illustrates the results for each test set $D_t$ when applying self-contact (gray background) and external-contact (yellow background) to both ChatGPT and LLaMA2-70B.
As observed earlier, applying self-contact, which reduces proficiency and intensifies preference bias, leads to an increased gap between $D_t$.
This substantial gap between $D_t$ indicates a decrease in robustness across various stages of emotional support, and in less proficient stages, they may provide poor-quality responses, which might worsen the seeker's situation and intensify distress.
In particular, all self-contact approaches significantly reduce performance on the exploration stage ($D_1$), which can create challenges in progressing to subsequent stages, ultimately hindering the achievement of the goals in emotional support.
On the other hand, external-contact reduces the overall gap between different $D_t$, particularly exhibiting significant improvement on ChatGPT.
This reduction contributes to robust performance in selecting strategy across the stages, which is crucial for effective emotional support.

\begin{figure}[t!]
\centering
    \begin{subfigure}[t]{0.96\linewidth}
        \includegraphics[width=\linewidth]{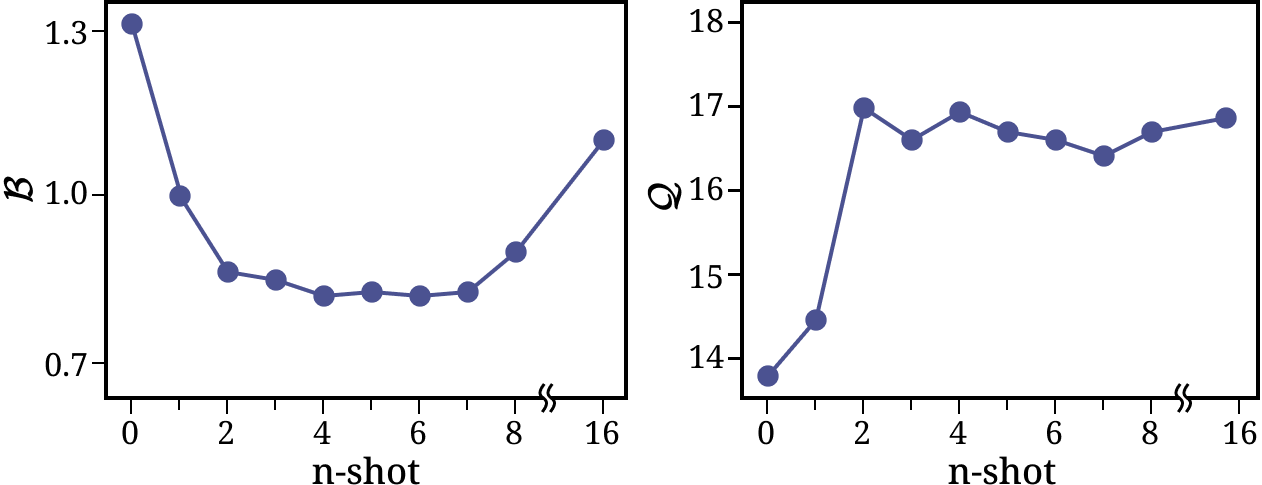} 
        \vspace{-0.7cm}
        \caption{}
        \label{fig:shot_results}
    \end{subfigure}
    \begin{subfigure}[t]{0.97\linewidth}
        \includegraphics[width=\linewidth]{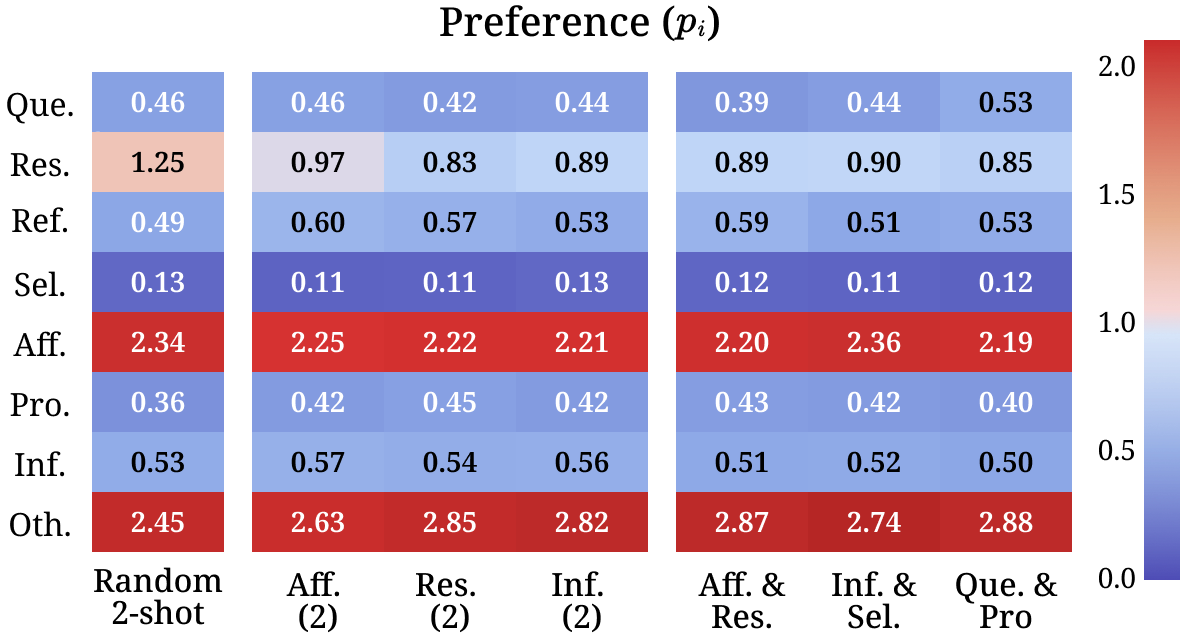}
        \caption{}   
        \label{fig:shot_controll_exp}
    \end{subfigure}
\caption{The results of (a) the variation in the number of shot examples, and (b) the effect of various combinations of strategies in 2-shot examples with ChatGPT.}
\label{fig:shot_total_results}
\end{figure}

\paragraph{Effect of examples in the prompt.}
\label{par:effect_of_shot}
To assess the efficacy of examples in the prompt, we initially investigate a trend associated with the number of examples ($n$).
Figure~\ref{fig:shot_results} demonstrates that proficiency and preference bias improve when using randomly selected examples.
However, while proficiency $\mathcal{Q}$ converges as $n$ increases, preference bias $\mathcal{B}$ worsens significantly with larger values of $n$ ($n>8$), indicating that too many examples may be detrimental.
Additionally, to understand the impact of different types of strategies employed in the examples, we include the various combinations of strategies within 2-shot examples.
Intriguingly, Figure~\ref{fig:shot_controll_exp} reveals consistent results across the diverse combinations.
In summary, providing the appropriate number of examples may enhance preference bias, whereas the type of strategies within each example does not matter.
Further analysis of each preference $p_i$ based on $n$ is in Appendix~\ref{app:few_shot_ablation}.

\begin{table}[t!]
\centering
\small
\resizebox{0.92\columnwidth}{!}{
\begin{tabular}{l c c cc ccc cccc c}
    \toprule
     &  &  & $D_1$ & $D_2$ & $D_3$  \\
    \cmidrule{4-6}
    \textbf{Base Models} & $\mathcal{Q}$ $\uparrow$ & $\mathcal{B}$ $\downarrow$ & \multicolumn{3}{c}{weighted-F1} \\
    \midrule
    BERT  & 18.02 & 0.50 & 18.17 & 22.68 & 19.25 \\
    RoBERTa  & 21.01 & 0.60 & 21.34 & 24.18 & 22.99 \\
    Mistral  & 21.89 & 0.45 & 22.61 & 23.57 & 24.59 \\
    \cmidrule{1-6}
    LLaMA2-7B  & 21.10 & 0.36 & 22.59 & 21.85 & 23.77 \\
     \bottomrule
\end{tabular}}
\caption{The results on the strategies selected by different strategy planners. Each model is fine-tuned with a uniform dataset across strategies.}

\label{tab:strategy_planner_backbone}
\end{table}

\paragraph{Various models as a strategy planner.}
In our previous experiments, a trained LLaMA2-7B serves as a strategy planner, yielding improved outcomes.
To explore the potential of various models as a strategy planner, we ablate with several language models, including Mistral and encoder-based models such as BERT~\citep{Devlin2019BERTPO} and RoBERTa~\citep{liu2019roberta}.
As shown in Table~\ref{tab:strategy_planner_backbone}, we find that using LLMs as the backbone model for the strategy planner leads to notable enhancements in proficiency and preference bias.
Moreover, while encoder-based models achieve performance comparable to LLMs, they exhibit relatively higher preference bias, indicating weaker robustness and potentially providing poor-quality emotional support.
We also leave the exploration of training a strategy planner with more diverse and systematic methods for future work.
Additionally, a more ablation study on directly fine-tuning LLMs as emotional supporters is provided in Appendix~\ref{app:fine_tuning_ablation}.

\subsection{RQ3: Does improving preference bias help to become a better emotional supporter?}
\label{sec:human_evaluation}

\paragraph{Criteria of human evaluation.}
To precisely assess whether responses provide helpful emotional support, we build a comprehensive set of criteria formulated in collaboration with psychologists in terms of emotional support, based on the perspective of \textbf{seeker's satisfaction} (\textit{Sat.}).
As emotional support aims to appropriately assess the user's state and reduce emotional intensity, we fine-grain this perspective and finally construct three smaller criteria to enable a more elaborate assessment:
(1) \textbf{Acceptance}: Does the seeker accept without discomfort;
(2) \textbf{Effectiveness}: Is it helpful in shifting negative emotions or attitudes towards a positive direction;
(3) \textbf{Sensitivity}: Does it take into consideration the general state of the seeker.
Furthermore, to clarify the capability of LLMs to align strategy and responses, we include \textbf{Alignment}.

We randomly sample 100 dialogues from three test sets ($D_t$), ensuring diversity (\eg, strategy), and three annotators are required to determine the \textit{Win/Tie/Lose} for each comparison in Table~\ref{tab:humaneval_results}.
Additionally, we ask three annotators to evaluate each sample on a 1-5 Likert scale, providing specific rubrics for each score to ensure detailed assessments on the quality of responses (Table~\ref{tab:poor_es}).
We include more details on the human evaluation, including the results of \textbf{Alignment}, in Appendix~\ref{app:es_eval}.

\begin{table}[t!]
\centering
\small
\resizebox{0.98\columnwidth}{!}
{%
\begin{tabular}{l ccc c}
    \toprule
    \textbf{ChatGPT} & Acc. & Eff. & Sen. & \textbf{Sat.}\\
    \midrule
     Vanilla  & 27.9 & 23.5 & 22.1 & 24.5\\
     Tie  & 20.6 & 32.4 & 22.1 & 25.0\\
     + Self-Refine & $\textbf{51.5}^\ddagger$ & $\textbf{44.1}^\ddagger$ & $\textbf{55.9}^\ddagger$ & $\textbf{50.5}^\ddagger$\\
     \midrule
     Vanilla & 22.9 & 24.0 & 14.6 & 20.5\\
     Tie & 21.9 & 33.3 & 27.1 & 27.4\\
     + w/ COMET  & $\textbf{55.2}^\ddagger$ & $\textbf{42.7}^\dagger$ & $\textbf{58.3}^\ddagger$ & $\textbf{52.1}^\ddagger$\\
    \midrule
     Vanilla & 13.1 & 25.3 & 16.2 & 18.2\\
     Tie & 26.3 & 26.3 & 21.2 & 24.6\\
     + w/ Example Expansion & $\textbf{60.6}^\ddagger$ & $\textbf{48.5}^\dagger$ & $\textbf{62.6}^\ddagger$ & $\textbf{57.2}^\ddagger$ \\
      \midrule
     Vanilla & 16.7 & 29.2 & 29.2 & 25.0 \\
     Tie & 12.5 & 16.7 & 12.5 & 13.9 \\
     + w/ Strategy Planner & $\textbf{70.8}^\ddagger$ & $\textbf{54.2}^\ddagger$ & $\textbf{58.3}^\ddagger$ & $\textbf{61.1}^\ddagger$ \\
    \bottomrule
\end{tabular}
}
\caption{The results of comparative human evaluation between various methods applied to ChatGPT and vanilla ChatGPT. ($\dagger$/$\ddagger$: p-value < 0.1/0.05 )}
\label{tab:humaneval_results}
\end{table}

\begin{table}[t!]
\centering
\small
\resizebox{0.92\columnwidth}{!}
{%
\begin{tabular}{l ccc}
    \toprule
    \textbf{Methods} & < 3 (\textit{fail}) & $\geqq$ 3 (\textit{acceptable})\\
    \midrule
    ChatGPT & \textbf{16.7} & 83.3 \\
    + Direct-Refine & \textbf{21.2} & 78.8\\
    + Self-Refine & \textbf{17.4} & 82.6\\
    + w/ Strategy planner & \textbf{8.0} & 92.0\\
    + Oracle Strategy & \textbf{3.8} & 96.2 \\
    \bottomrule
\end{tabular}
}
\caption{The ratio (\%) of scores below 3 (fail) and scores of 3 or above (acceptable) in Seeker's Satisfaction (\textit{Sat.}).}
\label{tab:poor_es}
\end{table}

\paragraph{Benefits of mitigating preference bias.}
Table~\ref{tab:humaneval_results} presents a comparative human evaluation between the results of various methods on ChatGPT and the results of vanilla ChatGPT.
Consistent with our previous findings, external-contact outperforms self-contact (\ie, Self-Refine) in terms of overall seeker's satisfaction (\textit{Sat.}).
Concretely, when comparing the \textit{w/ COMET} with \textit{Self-Refine}, which have similar proficiency but significant differences in preference bias, the overall seeker's satisfaction score is higher for \textit{w/ COMET} with lower preference bias.
Furthermore, among the external-contact methods, responses generated through the strategy planner, which exhibits the most significant improvements in preference bias, are the most helpful in reducing the seeker's emotional intensity.
Consequently, we can confirm that it is crucial to mitigate preference bias to enhance robustness in predicting strategy, thereby providing effective emotional support.

\paragraph{Drawbacks of aggravating preference bias.}
To understand the negative impact of severe preference bias, we investigate the proportion of responses that could worsen the seeker's situation or distress (\ie, rated below 3). 
Table~\ref{tab:poor_es} demonstrates that the proportion of poor-quality emotional support significantly increases in self-contact (\ie, Direct-Refine, Self-Refine), which exacerbates preference bias.
This confirms that the aggravation in preference bias sharpens the contrast between proficient and less proficient stages, leading to providing more poor-quality responses in the less proficient stages.
Additionally, the decrease in the proportion of poor-quality responses in external-contact (\ie, w/ Strategy Planner), where preference bias diminishes, supports this conclusion.
As a result, high preference bias disturbs robustness, leading to an increased number of poor-quality responses.
This demonstrates that low preference bias reduces the number of poor-quality responses and, consequently, is crucial for effective emotional support.

\section{Discussion and Conclusions}

This work conducts a strategy-centric analysis to delve into why LLMs struggle with providing emotional support, relying on the importance of strategy in emotional support.
Our results show that as LLMs exhibit preference bias towards certain strategies, they lack robustness in predicting strategy across the three stages of emotional support, where struggling in a particular stage may hinder the progress to the next stage.
We empirically demonstrate that LLMs are aligned with the psychological Contact Hypothesis just like humans, indicating that 
external assistance can mitigate the preference bias in LLMs, which they can not do themselves.
We highlight that mitigating the preference bias strengthens robustness in selecting appropriate strategies across the stages, leading to overall improvement in the quality of emotional support and a significant reduction in the number of poor-quality responses.
We hope that this work will become a promising step for future work to enhance the emotional intelligence of LLMs.

\section*{Limitations}\label{sec:limitations}
This work has the following limitations:
(1) 
As aforementioned in Section~\ref{sec:dataset}, \citet{Cheng2022MultiESC} demonstrate that the strategy \textit{Others} are not helpful in enhancing the response generation and may not be fully fine-grained.
This can potentially prevent obtaining sufficient insights by obscuring more detailed preferences of the model;
(2)
We include 2-shot examples for open-source LLMs as they often struggle to adhere to the desired output format (\eg, wrong strategy that is not among the eight provided).
Since we demonstrated improvement when prompting with n-shot examples in Section~\ref{par:effect_of_shot}, the actual proficiency and preference bias of open-source LLMs may be worse than the scores we published;
(3)
Understanding the reasons for preference bias is challenging not only for closed-source LLMs but also for open-source LLMs, as it is difficult to precisely grasp the relationships between strategy, training data, methods and model architecture;
(4)
We have observed that even when using an oracle strategy in LLMs (Table~\ref{tab:poor_es}), responses that increase emotional intensity still exist (3.8\%).
This indicates a lack of ability to generate appropriate responses for emotional support, even when the strategy is perfectly selected.
Therefore, future work might consider both correctly predicting the strategy and generating helpful responses based on the predicted strategy;
(5)
While we confirm that LLMs generally generate well-aligned responses with the strategy (Figure~\ref{fig:case_study_misalignment}), it is evident that there are some cases where they are not aligned, thereby future work should recognize this misalignment.

\section*{Ethical Considerations}
The ESConv a dataset used in this work is a publicly available and well-constructed benchmark for emotional support conversation, which is collected by employed crowd-sourced workers, with the sensitive and private information filtered during the dataset construction.
All participants in our human evaluation are volunteered, transparently informed of our research intent, and paid reasonable wages.

It is worth mentioning that the term "emotional support" in this paper mainly refers to support within a social context, such as interactions with friends or family in daily conversation, rather than professional counseling or diagnosis.
Moreover, as LLMs can generate sensual, harmful, biased, offensive, or violent content, using them as emotional support systems requires particular caution to avoid such content from appearing to users.
And it also requires considerable further efforts to construct a safer system, which is capable of detecting users who have tendencies of self-harming or suicide.

\section*{Acknowledgements}
This work was supported by Institute of Information \& Communications Technology Planning \& Evaluation (IITP) grant funded by the Korean government (MSIT)(No.RS-2020-II201361, Artificial Intelligence Graduate School Program (Yonsei University)) and (No.RS-2021-II212068, Artificial Intelligence Innovation Hub) and (No.RS-2022-II220077,AI Technology Development for Commonsense Extraction, Reasoning, and Inference from Heterogeneous Data). Jinyoung Yeo is The corresponding author.


\bibliography{anthology,custom}

\appendix

\begin{table*}[t!]
\centering
\small
\resizebox{0.95\textwidth}{!}{
\begin{tabular}{l c c c c c c c}
    \toprule
    & Ground-Truth & \multicolumn{2}{c}{GPT-4} & \multicolumn{2}{c}{ChatGPT} & \multicolumn{2}{c}{LLaMA2-70B} \\
    \cmidrule(lr){2-2}\cmidrule(lr){3-4}\cmidrule(lr){5-6}\cmidrule(lr){7-8}
    \textbf{Strategy} & ratio ($\%$) & ratio ($\%$) & preference & ratio ($\%$) & preference & ratio ($\%$) & preference \\
    \midrule
    Question & 16.6 & 1.4 & 0.11 & 1.4 & 0.12 & 19.6 & 1.50\\
    Restatement or Paraphrasing & 7.4 & 0.0 & 0.00 & 2.2 & 0.27 & 8.0 & 0.97\\
    Reflection of feelings & 12.0 & 10.2 & 0.92 & 14.4 & 1.48 & 11.0 & 0.85\\
    Self-disclosure & 12.9 & 4.0 & 0.26 & 2.0 & 0.14 & 7.3 & 0.48\\
    Affirmation and Reassurance & 17.9 & 60.0 & 4.26 & 64.0 & 4.49 & 32.0 & 1.88\\
    Providing Suggestions & 16.1 & 20.7 & 1.83 & 7.6 & 0.61 & 11.2 & 0.65\\
    Information & 11.9 & 2.8 & 0.34 & 6.6 & 0.59 & 6.2 & 0.48\\
    Others & 5.2 & 0.9 & 0.28 & 1.7 & 0.29 & 4.7 & 1.18\\
    \midrule
    Total & 100 & 100 & 8.00 & 100 & 8.00 & 100 & 8.00\\
     \bottomrule
\end{tabular}}
\caption{The ratio (\%) of strategy selected by LLMs and their preference ($p_i$) across the strategies.}

\label{tab:llm_esconv_preference}
\end{table*}

\section{Details of Preliminary Studies}

For the preliminary study, we prompt \textit{gpt-4-0613} and \textit{gpt-3.5-turbo-1106} to predict a strategy and generate a strategy-constrained response in 0-shot setting, and LLaMA2-7B in 2-shot setting as it struggles with adhering to desired output format.
We utilize a total of 4,833 samples across various strategies, and the strategy distribution of samples is reported in Table~\ref{tab:llm_esconv_preference} (Ground-Truth).
We provide the prompt used for the test in Table~\ref{tab:prompt_initial}.

\subsection{Analysis of LLMs on ESC}
\label{app:motivation_llms}

\paragraph{Performance in Selecting Correct Strategy.}
Table~\ref{tab:llm_esconv_proficiency} indicates that LLMs have limited proficiency in accurately predicting strategy, showing performance similar to random selection.

\begin{table}[h!]
\centering
\small
\resizebox{0.84\columnwidth}{!}{
\begin{tabular}{l c c }
    \toprule
    \textbf{Models} & accuracy (\%) & weighted-F1 \\
    \midrule
     \textit{random} & 12.6 & 13.0 \\
     \midrule
     GPT-4 & 22.1 & 17.5\\
     ChatGPT & 20.5 & 15.7\\
     LLaMA2-70B & 17.5 & 15.4\\
     \bottomrule
\end{tabular}}
\caption{The performance of strategy prediction for LLMs. The \textit{random} represents the results when strategies are randomly selected.}
\label{tab:llm_esconv_proficiency}
\end{table}

\paragraph{Preference for Strategy.}
To further analyze the reason behind the low performance, we investigate the distribution of how often LLMs select each strategy.
Table~\ref{tab:llm_esconv_preference} includes the proportions of strategy selected by LLMs and their preferences ($p_i$) for each strategy.
We have observed that all LLMs have a strong preference for the strategy \textit{Affirmation and Reassurance} and each LLM has its preferred strategies with various degrees of preference.

\subsection{Importance of Strategy}
\label{app:importance_of_strategy}

To comprehend the importance of strategy in emotional support conversation tasks using LLMs, we examine \textit{gpt-3.5-turbo-1106} and LLaMA2-70B under the following settings: response generation (a) without strategy, (b) with randomly selected strategy, (c) with strategy predicted by itself, and (d) with ground-truth strategy.

Figure~\ref{fig:strategy_importance} and Table~\ref{tab:strategy_exp_app} represent that the responses based on correct strategy (\textit{ground-truth strategy}) outperforms those generated without strategy.
Furthermore, although LLMs exhibit low performance in strategy prediction, the responses conditioned on predicted strategy achieve performance similar to those without strategy, emphasizing that there is significant room for improvement in the quality of emotional support responses with LLMs.

\begin{table}[h!]
\centering
\small
\resizebox{0.96\columnwidth}{!}
{%
\begin{tabular}{l c c ccc c ccc}
    \toprule
    \textbf{Models} & \textbf{Strategy} & $\mathcal{Q}$ & \textbf{R-L.} & \textbf{Sat.} \\ 
    \midrule
    \multirow{4}{*}{
    \begin{tabular}{@{}l@{}}
          \textbf{ChatGPT}
    \end{tabular}} & no & - &  \underline{15.25} & 3.94 \\
     & random & 12.21 & 14.90 & 3.92 \\ 
     & predicted & 15.04 & 15.19 & \underline{4.00} \\
     \cmidrule(lr){2-10}
     & Ground-truth & - & \textbf{17.16} & \textbf{4.06}\\
     \midrule
    \multirow{4}{*}{
    \begin{tabular}{@{}l@{}}
          \textbf{LLaMA2 (70B)}
    \end{tabular}} & no & - & \underline{14.92} & 3.80\\
     & random & 12.21 & 14.10 & 3.87 \\
     & predicted & 14.55 & 14.66 & \underline{3.89} \\
     \cmidrule(lr){2-10}
     & Ground-truth & - & \textbf{17.13} & \textbf{4.02}\\ 
    \bottomrule
\end{tabular}}
\caption{The results of both automated and human evaluation for the responses from ChatGPT and LLaMA2-70B. The responses are generated with/without strategy. The best results are \textbf{bolded} and the second best are \underline{underlined}.}
\label{tab:strategy_exp_app}
\end{table}

\section{ESConv Dataset}
\label{app:ESConv_detail}

\subsection{Definitions of Stages}
\label{app:stage_and_strategy_detail}

Grounded on Hill's Helping Skills Theory~\citep{Hill2009Helping}, \citet{Liu2021Towards} propose three stages of emotional support:
\begin{enumerate}
    \item \textbf{Exploration}: Explore to identify the seeker's problem.
    \item \textbf{Comforting}: Comfort the seeker through expressing empathy and understanding.
    \item \textbf{Action}: Help the seeker solve the problems.
\end{enumerate}
Although it is suggested that ESC target these stages in the order: (1) Exploration $\rightarrow$ (2) Comforting $\rightarrow$ (3) Action, this sequence can be flexibly tailored to individual needs, as conversations, in practice, do not always follow a fixed order.

\subsection{Definitions of Strategies}
\citet{Liu2021Towards} also propose a specific set of conversational skills corresponding to each stage.
In ESConv, they annotate eight types of support strategies:
\begin{itemize}[noitemsep]
    \item \textbf{Question}: Asking for information related to the problem to help the seeker articulate the issues that they face.
    \item \textbf{Restatement or Paraphrasing}: A simple, more concise rephrasing of the seeker's statements that could help them see their situation more clearly.
    \item \textbf{Reflection of Feelings}: Articulate and describe the seeker's feelings to show an understanding of the situation and empathy.
    \item \textbf{Self-disclosure}: Divulge similar experiences that you have had or emotions that you share with the help-seeker to express your empathy. 
    \item \textbf{Affirmation and Reassurance}: Affirm the seeker's ideas, motivation, strengths, and capabilities to provide reassurance and encouragement. 
    \item \textbf{Providing Suggestions}: Provide suggestions about how to get over the tough and change the current situation, but be careful to not overstep and tell them what to do. 
    \item \textbf{Information}: Provide useful information to the help-seeker, for example with data, facts, opinions, and resources.
    \item \textbf{Others}: Use other support strategies that do not fall into the above categories. 
\end{itemize}

\label{app:dataset_statistics}

\section{Experiments Details}

\subsection{Evaluation Sets}

In this study, we systematically partition the ESConv dataset into three distinct test sets, denoted as $D_1$ (Exploration), $D_2$ (Comforting), and $D_3$ (Action), to facilitate stage-specific assessments.
To prevent utterance duplication, we split the 1,300 dialogues within the ESConv dataset into three sets and randomly allocate them to $D_t$. 
We slice each dialogue comprising 5 to 15 turns to generate instances.
The determination of the stage for the label response of each instance is based on the majority stage indicated by surrounding strategies within a window size of 4.
In cases where the randomly assigned stage of $D_t$ differs from the determined stage, the instance is excluded from the respective test set.
Furthermore, to maintain the relevance of the test sets to emotional support contexts, we restrict the slicing process, ensuring that the frequency of the \textit{Others} strategy does not exceed 5\%.
Detailed statistics of the test sets are provided in Table~\ref{tab:test_set_statistics} and Table~\ref{tab:dataset_statistics}.

\subsection{Preference Metric}
\label{app:evaluation_setup}

\paragraph{Bradley-Terry Model.}
\label{app:bt_modeling}
The Bradley-Terry model (BT model) serves as a probability model for pairwise comparisons between individuals or objects.
Its utility spans a broad spectrum of areas, notably in ranking competitors in sports, chess, and other competitions. 
Beyond these traditional domains, the BT model extends to the realm of machine learning, facilitating multi-class probability estimations by incorporating pairwise classification results.
Recently,~\citet{rafailov2023direct} employed the BT model for optimizing preference alignment of LLMs, known as direct preference optimization.

\paragraph{Preference Evaluation with the Bradley-Terry Model.}
In this study, we employ BT modeling to assess the preference of LLMs across the strategies.
The probability $P(i>j)$, representing the preference for strategy $i$ over ground-truth strategy $j$, is formally defined as:
\begin{equation}
    P(i>j)=\frac{p_{i}}{p_{i} + p_{j}}
\end{equation}
where we assign a numerical score $s_i$ to each strategy $i$ and define $p_i=e^{s_i}$, enabling the expression of $P(i>j)$ in terms of these scores.
~\citet{zermelo1929zermelo_algorithm} characterizes the parameter $p_i$ as \textit{playing strengths}.
In scenarios involving a series of pairwise competitions among \textit{N} competitors (specifically, 8 strategies in our case), estimating these strengths becomes relatively straightforward. 

The likelihood of the preference ($\mathbf{P}$) with the Bradley-Terry model is given by the equation:
\begin{equation}
\notag
    \mathbf{P} = \prod_{ij}{[P(i>j)]}^{w_{ij}} = \prod_{ij}\Big(\frac{p_i}{p_i + p_j}\Big)^{w_{ij}} 
\end{equation}
where $w_{ij}$ represents the total number of times where strategy $i$ is preferred over strategy $j$.
This leads to the log-likelihood:
\begin{align}
\notag
    \log{\mathbf{P}} & = \sum_{ij}{w_{ij}}\log{\frac{p_i}{p_i + p_j}} \\
    \notag
    & = \sum_{ij}{w_{ij}}\log{p_i} - \sum_{ij}{w_{ij}}\log{(p_i+p_j)}
\end{align}

~\citet{zermelo1929zermelo_algorithm} showed that this expression has only a single maximum, differentiating with respect to $p_i$ for any $i$ and setting the result to zero:
\begin{equation}
    \frac{1}{p_i}\sum_{j}{w_{ij}} - \sum_{j}{\frac{w_{ij}+w_{ji}}{p_i + p_j}} = 0
\label{eq:zermelo_eq}
\end{equation}


\paragraph{Iterative Algorithms.}
Following the efficient algorithm proposed by \citet{Newman2023Efficient_BT}, Eq~\ref{eq:zermelo_eq} can be rearranged as:
\begin{equation}
    \frac{1}{p_i}\sum_{j}{w_{ij}}\frac{p_j}{p_i + p_j}- \sum_{j}{\frac{w_{ji}}{p_i + p_j}} = 0
\end{equation}
\begin{equation}
    p_{i} =  \frac{\sum_{j}(w_{ij}p_{j})/(p_{i}+p_{j})}{\sum_{j}w_{ji}/(p_{i}+p_{j})}
\label{eq:BT_equation_app}
\end{equation}
Finally, Eq~\ref{eq:BT_equation_app} results in the iterative algorithm for the Bradley-Terry model to calculate the preference $p_i$ for each strategy $i$.


For this iterative algorithm, we initially set all values ($p_i$) to 1 and iteratively update these estimates over $k$ iterations, where in this study we utilize 20 iterations for estimation.
Subsequent to each iteration, it is necessary to normalize the values by dividing them by their geometric mean to ensure stability and convergence of the algorithm. 
This normalization step is represented as:
\begin{equation}
\normalsize
    p_{i} \leftarrow \frac{p_{i}'}{\left( \Pi_{j=1}p_{j}'\right) ^{1/N}} 
\end{equation}
where $N$ is the total number of strategies.
After the final iteration, the converged $p$ values indicate the final preference $p_i$ for strategy $i$.

\begin{table}[h!]
\centering
\small
\resizebox{0.98\columnwidth}{!}
{\begin{tabular}{l ccc}
    \toprule
    \textbf{Category} & $D_1$ & $D_2$ & $D_3$ \\
    \midrule
    stage & Exploration & Comforting & Action \\
    \# of samples & 549 & 524 & 816 \\
    \# of dialogues & 433 & 434 & 433 \\
    Avg. \# of turns & 9.95 & 10.04 & 10.66 \\
    Avg. length of utterance & 16.27 & 16.81 & 18.92 \\
    \bottomrule
\end{tabular}}
\caption{Statistics of the processed ESConv dataset for our analysis.}
\label{tab:dataset_statistics}
\end{table}

\subsection{Models}
\label{app:model_implementation}

\paragraph{ChatGPT / GPT-4.}
ChatGPT and GPT-4~\citep{openai2023chatgpt, openai2023gpt4} are among the most widely used LLMs, demonstrating state-of-the-art performance in numerous applications.
However, as they are closed-source LLMs, they are available exclusively through APIs.
Thereby, we employ \textit{gpt-3.5-turbo-1106} for ChatGPT and \textit{gpt-4-0613} for GPT-4 in this work.

\paragraph{LLaMA2.} 
LLaMA2~\citep{Touvron2023Llama2} is a prestigious open-source LLM that is widely employed as a foundation model for various open-source LLMs. 
The model size ranges from 7B to 70B parameters.
In this work, we implement both the 7B (\textit{Llama-2-7b-hf}) and the 70B (\textit{Llama-2-70B-hf}) versions, allowing for an exploration of the effects of model size on performance.

\paragraph{Tulu.}
Tulu is a model with 70B parameters, based on LLaMA2 models fine-tuned on V2 mixture~\citep{ivison2023tulu}.
The employ the \textit{tulu-2-70b} version in our experiments to assess its capabilities within the context of our study.

\paragraph{Vicuna.} 
Vicuna is a 13B language model from LLaMA-13B model fine-tuned with high-quality conversation datas~\citep{zheng2023vicuna}.
We incorporate the \textit{vicuna-13b-v1.5} version into our experiments to evaluate its performance.

\paragraph{Solar.}
Solar is an LLM with 10.7B parameters, employing the depth up-scaling (DUS) method as its scaling method~\citep{kim2023solar}. 
This approach contributes to its performance exceeding other LLMs, including those utilizing mixture-of-experts (MoE) methods.
We use the \textit{SOLAR-10.7B-Instruct-v1.0} version in this work.

\paragraph{Mistral.}
Mistral is a 7B LLM that leverages grouped-query attention (GQA) and sliding window attention (SWA) for faster inference and reduced inference cost~\citep{jiang2023mistral}.
It claims superior performance over the LLaMA2-13B model and even the LLaMA-34B model across various evaluation benchmarks.
We employ the \textit{Mistral-7b-Instruct-v0.2} version.

\subsection{Prompts Details}
\label{app:prompt_details}

The prompts employed in our experiments are shown in Table~\ref{tab:prompt_initial}. 
To ensure a clear understanding of the task, \textit{Task description} and \textit{strategy description} are prompted to LLMs.
Furthermore, in addition to the \textit{dialogue context}, we also incorporate \textit{dialogue background}, which encompasses the seeker's problem, emotion, and situation gathered from a pre-chat survey.
Depending on the method employed, various types of information, such as feedback, rationale, commonsense knowledge, and few-shot examples, are also included as supplementary inputs.

\paragraph{Random few-shot samples.}
To prevent potential biases in strategy induced by few-shot learning, we randomly select examples.
During the experiments, for each data instance, we randomly select exemplars with non-overlapping strategies and incorporate them into the prompt.
This approach ensures that the influence of few-shot samples on strategy prediction is minimized by diversifying the strategies presented to the model.
However, we figure out in Section~\ref{sec:automatic_evaluation} and Figure~\ref{fig:shot_controll_exp} that the types of strategies included in the prompt as examples do not significantly impact on the results in the end.

\subsection{Methods Details}
\label{app:method_details}

\paragraph{Direct Refine.} 
Direct refine is a straightforward refinement method, wherein we instruct the model to revise its initial response to incorporate emotional support elements.

\paragraph{Self-Refine.} 
Self-refine, a method introduced by \citet{Madaan2023SelfRefine}, initiates by generating feedback emphasizing emotional support from the initial response. 
Subsequently, it refines the response based on this feedback.

\paragraph{Emotional-CoT.}
Building upon the success of Chain-of-Thought (CoT) prompting \citep{wei2022chain}, we employ CoT to first generate the \textit{user state}, which then guides the generation of strategy and response.

\paragraph{w/ COMET.}
To incorporate external commonsense knowledge for providing emotional support, we integrate the COMET model~\citep{Hwang2020COMETATOMIC2O}, specifically COMET-BART\footnote{https://github.com/allenai/comet-atomic-2020}, while leveraging five relation types (\ie, xReact, xIntent, xNeed, xEffect, and xWant).
Following~\citet{chae2023dialogue}, we implemented a retriever using ComFact~\citep{gao-etal-2022-comfact} to align the dialogues with the knowledge from COMET.
Among the inferences generated by COMET, we apply the retriever (DeBERTa-large\footnote{https://github.com/silin159/comfact}) and filter inferences that are non-relevant to the dialogue context.
Subsequently, we convert the remaining inferences into natural language and augment to LLMs, which is shown in Table~\ref{tab:prompt_initial}.

\paragraph{w/ Strategy Planner.}
Strategy planner is a classification model that is fine-tuned to predict the strategy based on dialogue background and context. 
Thereby, we formulate \textit{w/ Strategy Planner} as follows: given the dialogue background $\mathcal{I}$, and dialogue context $\mathcal{C}$, the strategy planner model $\theta'$ predicts the strategy $\mathcal{\hat{S}}$. Then, LLM $\theta$ generates the response $\mathcal{R}$, leveraging $\mathcal{I}$, $\mathcal{C}$, and $\mathcal{\hat{S}}$.
\begin{align}
\label{eq:strategyPlanner_formulation}
    \mathcal{\hat{S}} \sim P_{\theta'}(\cdot|\mathcal{I}, \mathcal{C}) \\
    \mathcal{R} \sim P_{\theta}(\cdot|\mathcal{I}, \mathcal{C}, \mathcal{\hat{S}} )
\end{align}

\section{Implementation Details}
\label{app:sft_details}

All experiments are conducted on 8 NVIDIA GeForce RTX 3090 GPUs and 2 NVIDIA A100 80GB PCIe GPUs.

\paragraph{Fine-tuning.}
Since the test sets are constructed by dividing the dialogues in ESConv into three without overlap, to evaluate each test set with a trained model, we construct a train/valid set from dialogues corresponding to the other two sets and train the model on it.

For training, we employ QLoRA \citep{dettmers2023qlora} to effectively fine-tune a model, incorporating 4-bit quantization and specifying the dimension of low-rank metrices as 64 and alpha as 16.
The DeepSpeed library\footnote{https://www.deepspeed.ai} is utilized to facilitate the training, with a learning rate of 5e-5 over 5 epochs, resulting in approximately 8 hours of training.
For encoder-based models like BERT and RoBERTa, we train them to classify among 8 categories (corresponding to the number of strategies), with training extending up to a maximum of 20 epochs.

\paragraph{Inference.}
For generating responses, we follow the default settings provided by OpenAI for top-$p$ sampling and temperature, with $p = 1.0$ and $T = 0.7$.
To achieve higher throughput during inference, we leverage the vLLM library\footnote{https://docs.vllm.ai}.

\paragraph{Terms and License.} 
For our implementation and evaluation, we use Huggingface library\footnote{\url{https://huggingface.co/}} and vLLM library. 
Both libraries are licensed under Apache License, Version 2.0.
We have confirmed that all of the artifacts used in this paper are available for non-commercial scientific use.

\section{Details on Human Evaluation}
\label{app:es_eval}

\subsection{Human Evaluation Criteria}

With automatic metrics, it is challenging to precisely assess the emotional support quality of responses~\citep{mehri2020usr, gao2022reamsharp}.
Furthermore, conventional criteria commonly used for general dialogue are not specifically designed to evaluate whether a response provides emotional support.
Hence, in collaboration with four psychologists, we develop a specific set of criteria focused on assessing whether a response provide effective emotional support from various perspectives of the seeker.

\textbf{Seeker's Satisfaction (\textit{Sat.})}, focusing on the quality of emotional support, comprises three detailed criteria.
Moreover, we add \textbf{Alignment} to assess how well the generated response aligns with the predicted strategy.
Consequently, we focus on these four criteria:

\begin{itemize}
  \item \textbf{Acceptance}: Is the response accepted by the seeker without discomfort or resistance?
  \item \textbf{Effectiveness}: Is it expected that the response would mitigate or shift the seeker's negative emotional state or attitude toward a more positive direction?
  \item \textbf{Sensitivity}: Does the response take into consideration the seeker's state (mood, needs, resources, culture, attitude, etc.)?
  \item \textbf{Alignment}: Is the response fitting for the chosen strategy?
\end{itemize}

\subsection{Implementations of Human Evaluation}
\begin{figure*}[t!]
    \centering
    \includegraphics[width=0.95\linewidth]{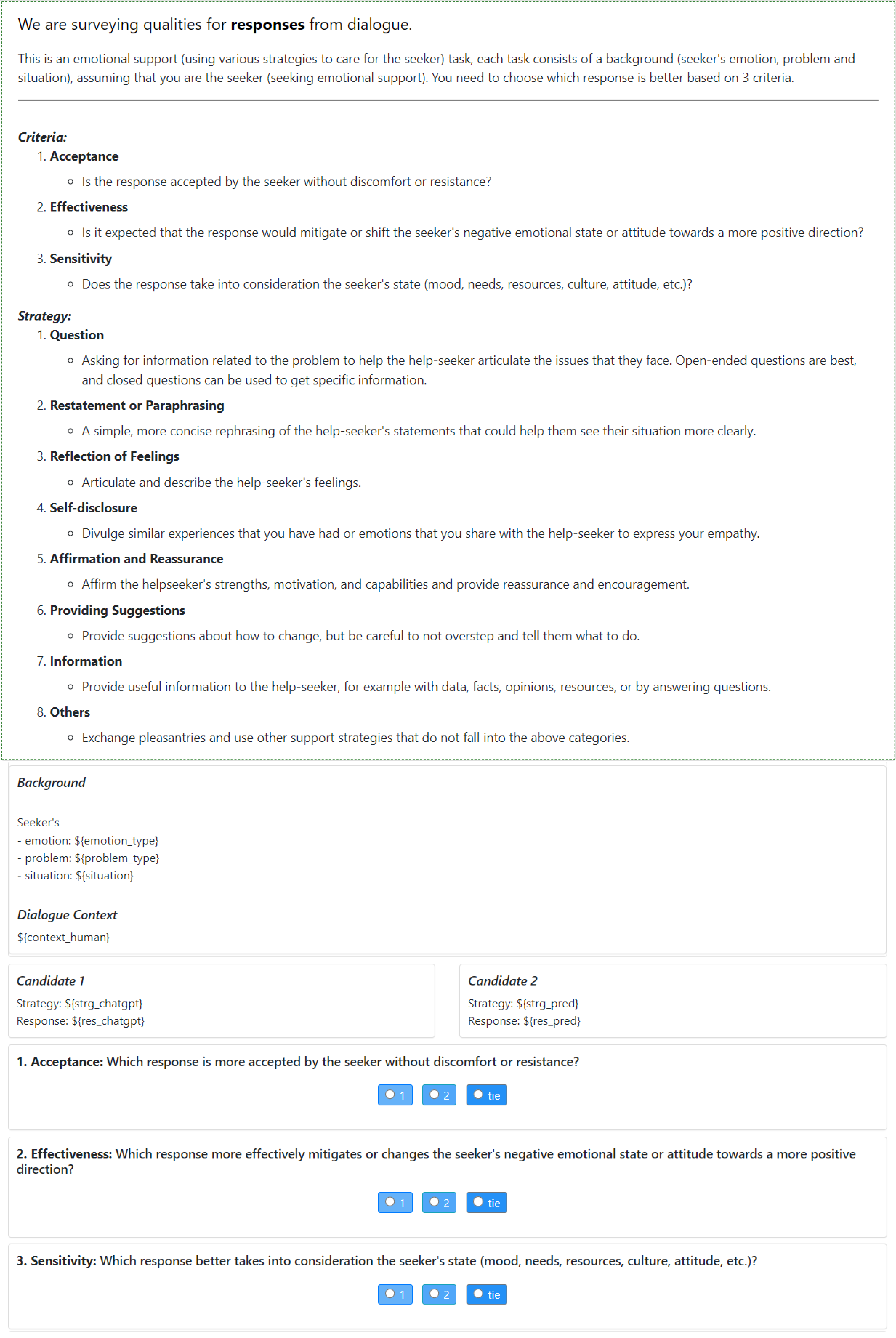}
    \caption{Interface for comparative human evaluation on \textbf{Seeker's Satisfaction} (\textit{Sat.}).}
\label{fig:interface_score_wl}
\end{figure*}

\begin{figure*}[t!]
    \centering
    \includegraphics[width=0.95\linewidth]{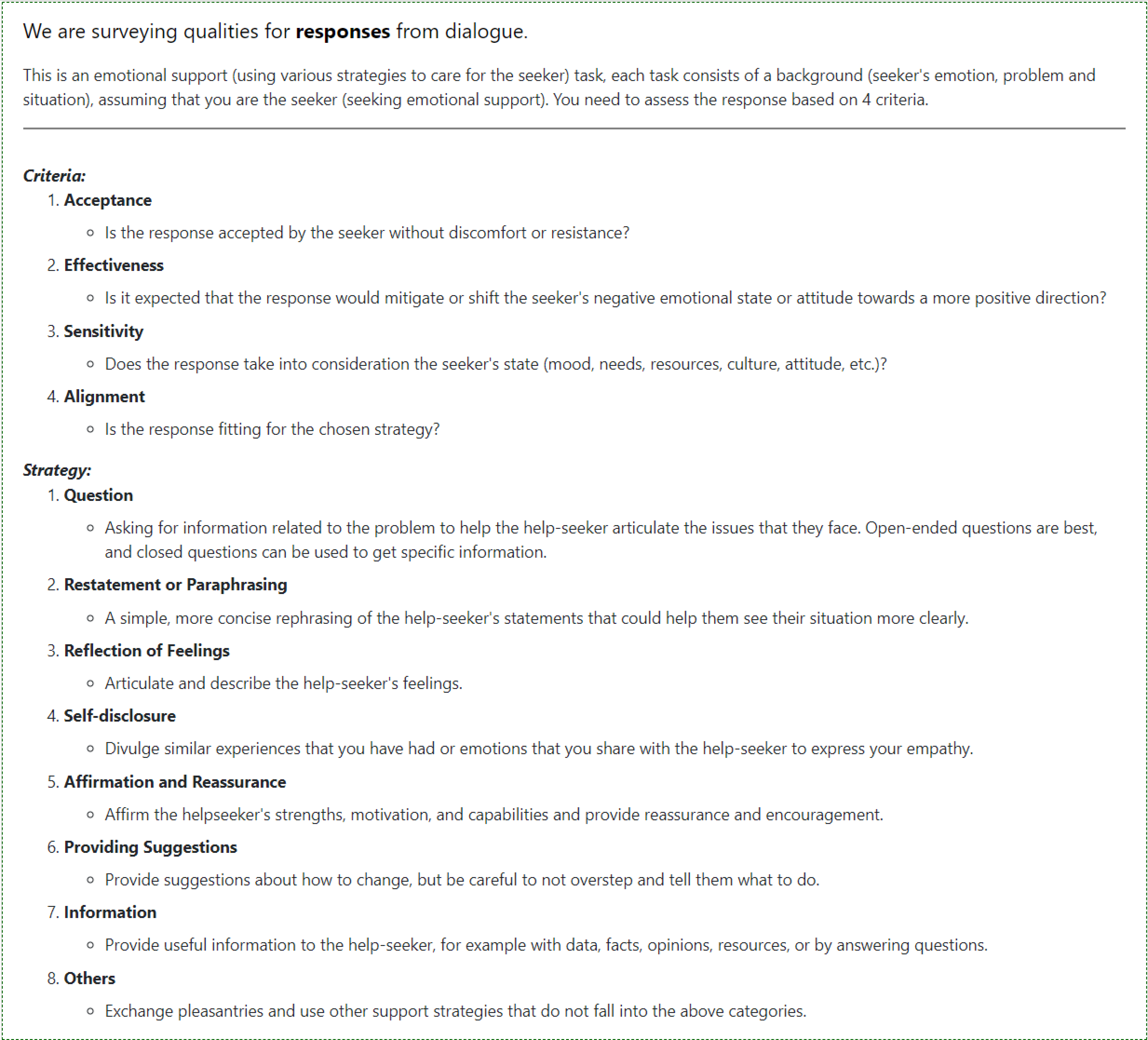}
    \caption{Interface for human evaluation on \textbf{Seeker's Satisfaction} (\textit{Sat.}) using 5-point Likert scale (Instruction part).}
\label{fig:interface_score_1}
\end{figure*}

\begin{figure*}[t!]
    \centering
    \includegraphics[width=0.95\linewidth]{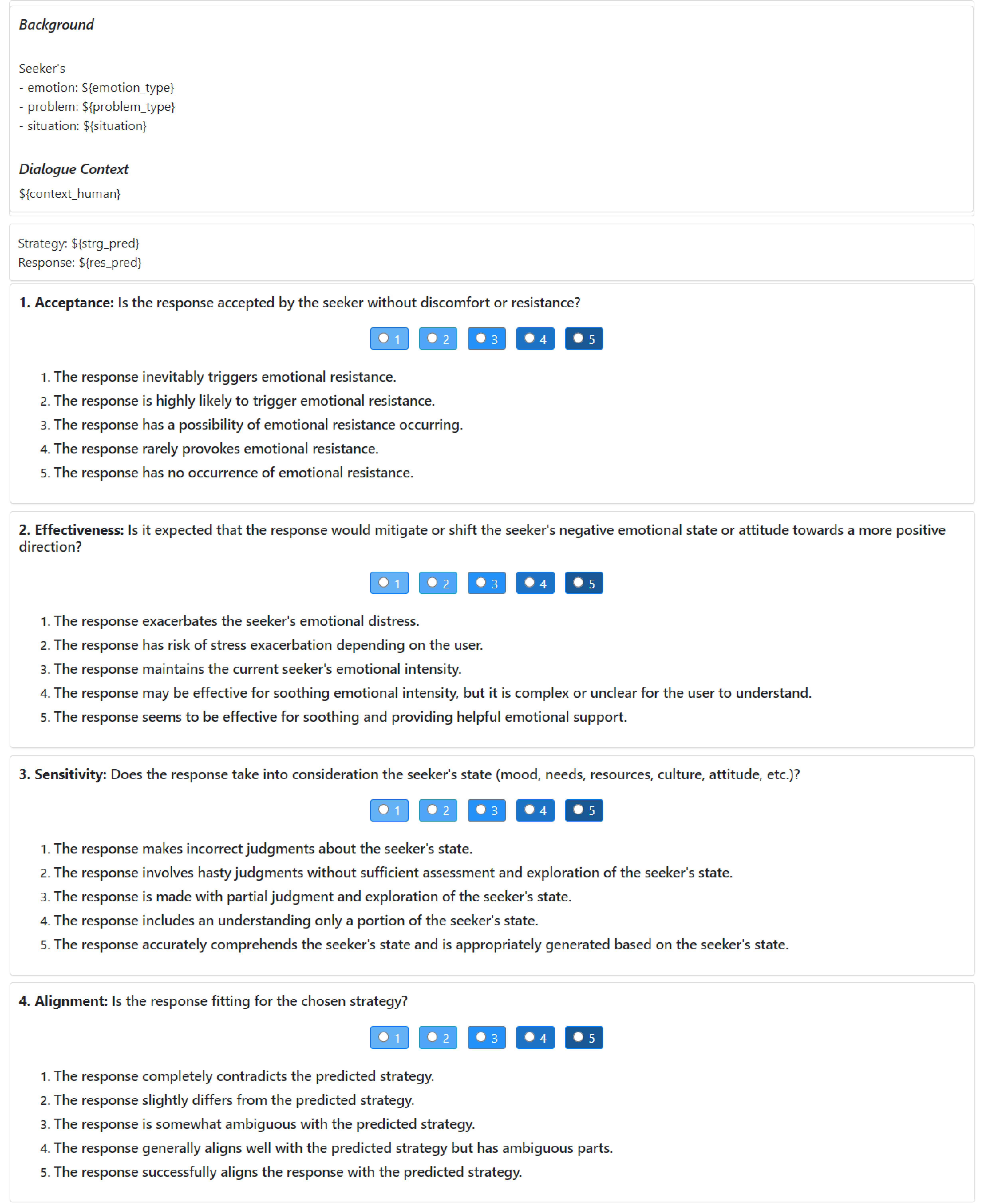}
    \caption{Interface for human evaluation on \textbf{Seeker's Satisfaction} (\textit{Sat.}) using 5-point Likert scale (Evaluation part).}
\label{fig:interface_score_2}
\end{figure*}

We employ human evaluation, outsourcing the task to assess response quality on Amazon Mechanical Turk (AMT).
Figure~\ref{fig:interface_score_wl} shows the interface employed for comparative evaluations (\textit{Win/Lose/Tie}) between two responses.
Figure~\ref{fig:interface_score_1} and~\ref{fig:interface_score_2} depict the interface employed to rate our four criteria using 5-point Likert scale.
Detailed instructions and rubrics for each score are included to ensure precise evaluation.
For each evaluation, we ask three human annotator to assess 100 samples each based on four specified criteria.
We compensate each data piece in the human evaluation with a payment of \$0.07.

\section{Additional Analysis}
\label{app:additional_analysis_llms}

\subsection{LLMs' Proficiency for Each Strategy}
\label{app:sec4_detailed_results}

\begin{figure*}[t!]
    \centering
    \includegraphics[width=0.88\linewidth]{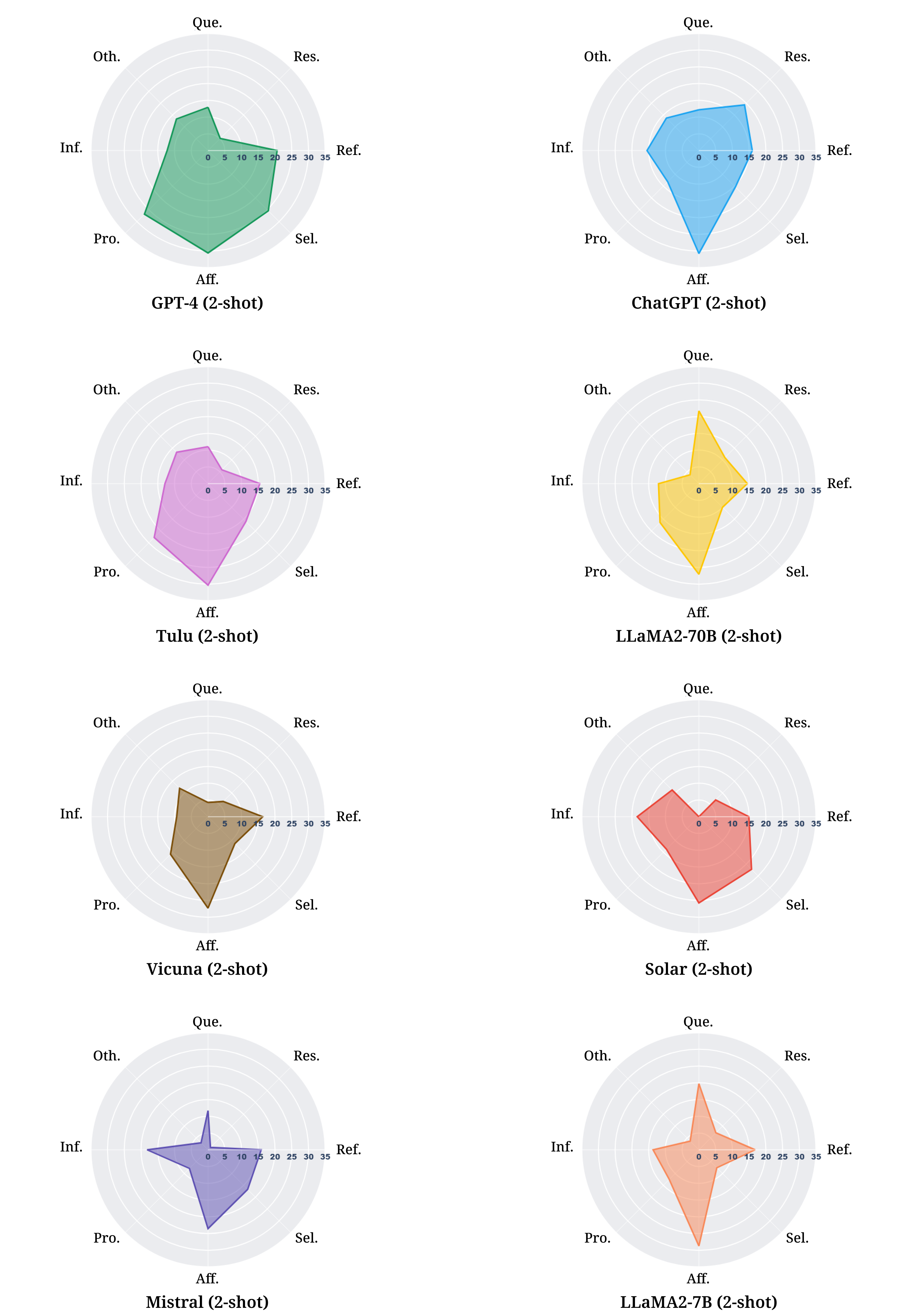}
    \caption{The proficiency by strategy (F1-score) on LLMs.}
\label{fig:llms_proficiency}
\end{figure*}

Building upon the findings where LLMs generally tend to demonstrate a low proficiency, as shown in Figure~\ref{fig:llms_strategy_results}, we further delve into the proficiency of each strategy on LLMs.
As illustrated in Figure~\ref{fig:llms_proficiency}, there are notable differences in proficiency depending on the strategy. 
In particular, each LLM tends to exhibit higher proficiency in strategies with higher preference, observed in Figure~\ref{fig:preference_figure}.

\begin{table}[h!]
\centering
\small
\resizebox{0.75\columnwidth}{!}
{
\begin{tabular}{lcc}
    \toprule
    \textbf{Models} & \textbf{Params} & \textbf{Pearson Correlation} \\
    \midrule
    GPT4 & - & 0.820 \\
    ChatGPT & 175B & 0.752 \\
    \midrule
    Tulu & 70B & 0.899 \\
    LLaMA2 & 70B & 0.772 \\
    Vicuna & 13B & 0.935 \\
    Solar & 10.3B & 0.747 \\
    Mistral & 7B & 0.943 \\
    LLaMA2 & 7B & 0.600 \\
     \bottomrule
\end{tabular}}
\caption{Relationship between preference and proficiency. The Pearson correlation between preference ($p_i$) and proficiency ($q_i$) of each strategy for LLMs.}

\label{tab:relationship_prefer_prof}
\end{table}

\subsection{Relation between Proficiency and Preference}
\label{app:relationship_pref_prof}

In Figure~\ref{fig:llm_preference_proficiency}, we observe that LLMs achieve higher scores on test sets aligned with strategies that they prefer more, raising the question of how this preference influences the proficiency.
To explore the relationship between preference $p_i$ and proficiency $q_i$, we calculate the Pearson correlation between $p_i$ and $q_i$ for each strategy.
As a result, Table~\ref{tab:relationship_prefer_prof} reports a strong positive correlation between preference and proficiency for most LLMs, suggesting that a high preference $p_i$ for strategy $i$ leads to a high proficiency $q_i$.
Ultimately, this confirms that LLMs perform better in stages containing preferred strategies.

\subsection{Preference for Strategies by the Number of Examples.}
\label{app:few_shot_ablation}

\begin{figure}[t!]
\centering
    \includegraphics[width=0.9\linewidth]{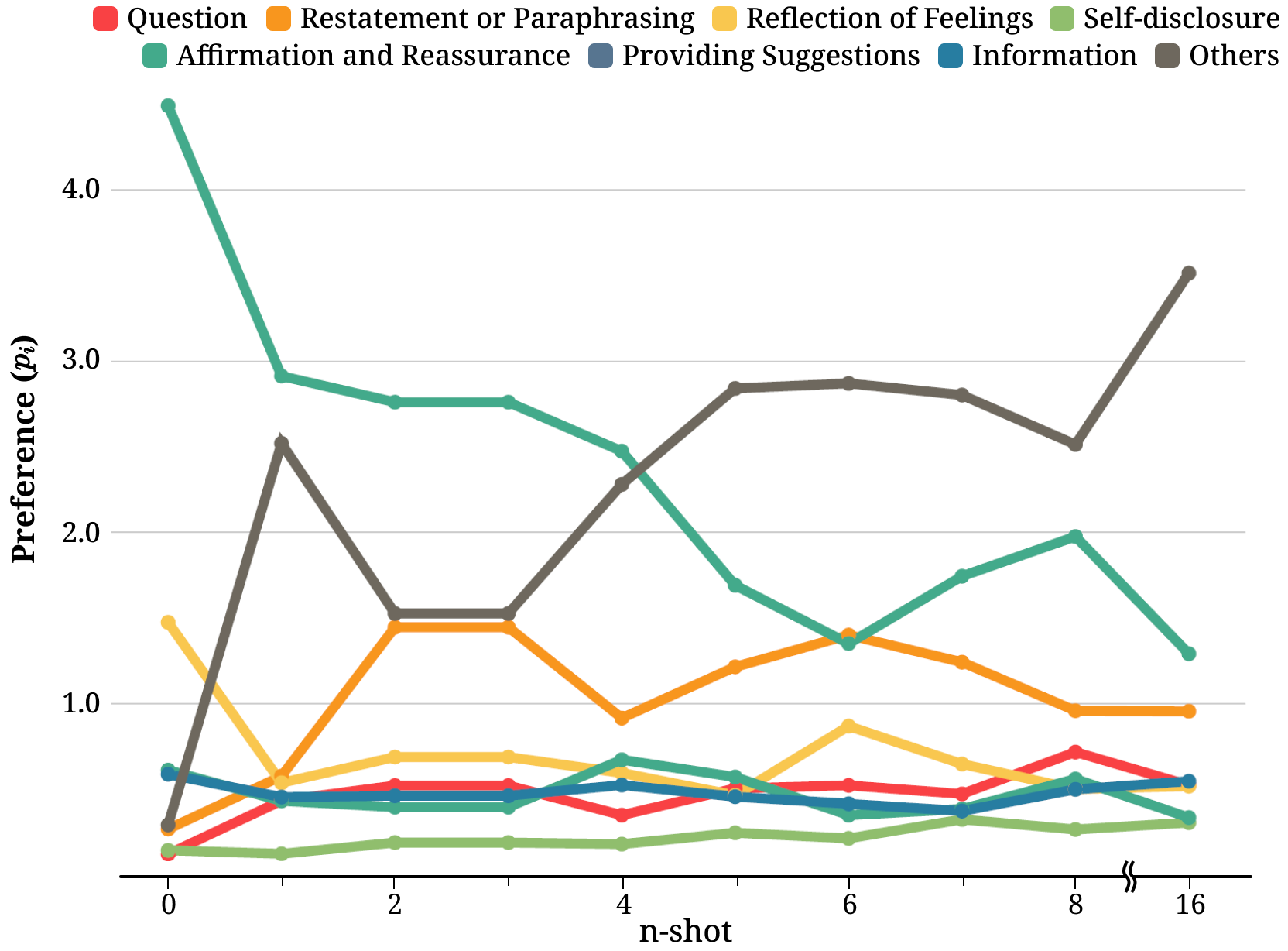}
\caption{The results of strategy preference as the number of shots increases.}
\label{fig:shot_controll}
\end{figure}

In Figure~\ref{fig:shot_results}, we observed improvements in proficiency and preference bias when prompting ChatGPT with few examples.
However, we also found that as the number of examples increases, preference bias significantly worsens.
To delve deeper into the reasons behind this, we examine the changes in preference for each strategy as the number of examples increases.
As demonstrated in Figure~\ref{fig:shot_controll}, the preference for \textit{Affirmation and Reassurance} gradually diminishes, while the preference for \textit{Others} gradually increases.
The strong preference for the \textit{Others}, as the number of examples increases, eventually exacerbates preference bias.
Consequently, the strong preference for the \textit{Others} disrupts the selection of alternative strategies, hindering the enhancement of proficiency as the number of shot examples increases.

\subsection{Supervised Fine-tuning on ESC Task}
\label{app:fine_tuning_ablation}

To explore the possibility of fine-tuning the model itself as an emotional supporter in addition to fine-tuning the strategy planner, we train the LLaMA2-7B to generate emotional support responses. 
Table~\ref{tab:training_results} shows that fine-tuning the model leads to significant improvements in emotional support quality.

We also ablate to examine the effectiveness of strategy on fine-tuned models.
As a result, Table~\ref{tab:training_results} demonstrates that fine-tuning the model on a dataset with strategies yields a higher quality of emotional support compared to training on a dataset that does not include strategies.

\section{Case Study}
\label{app:case_study_llm}

\subsection{Responses of LLMs by Stages}
\begin{figure*}[t!]
    \centering
    \includegraphics[width=0.95\linewidth]{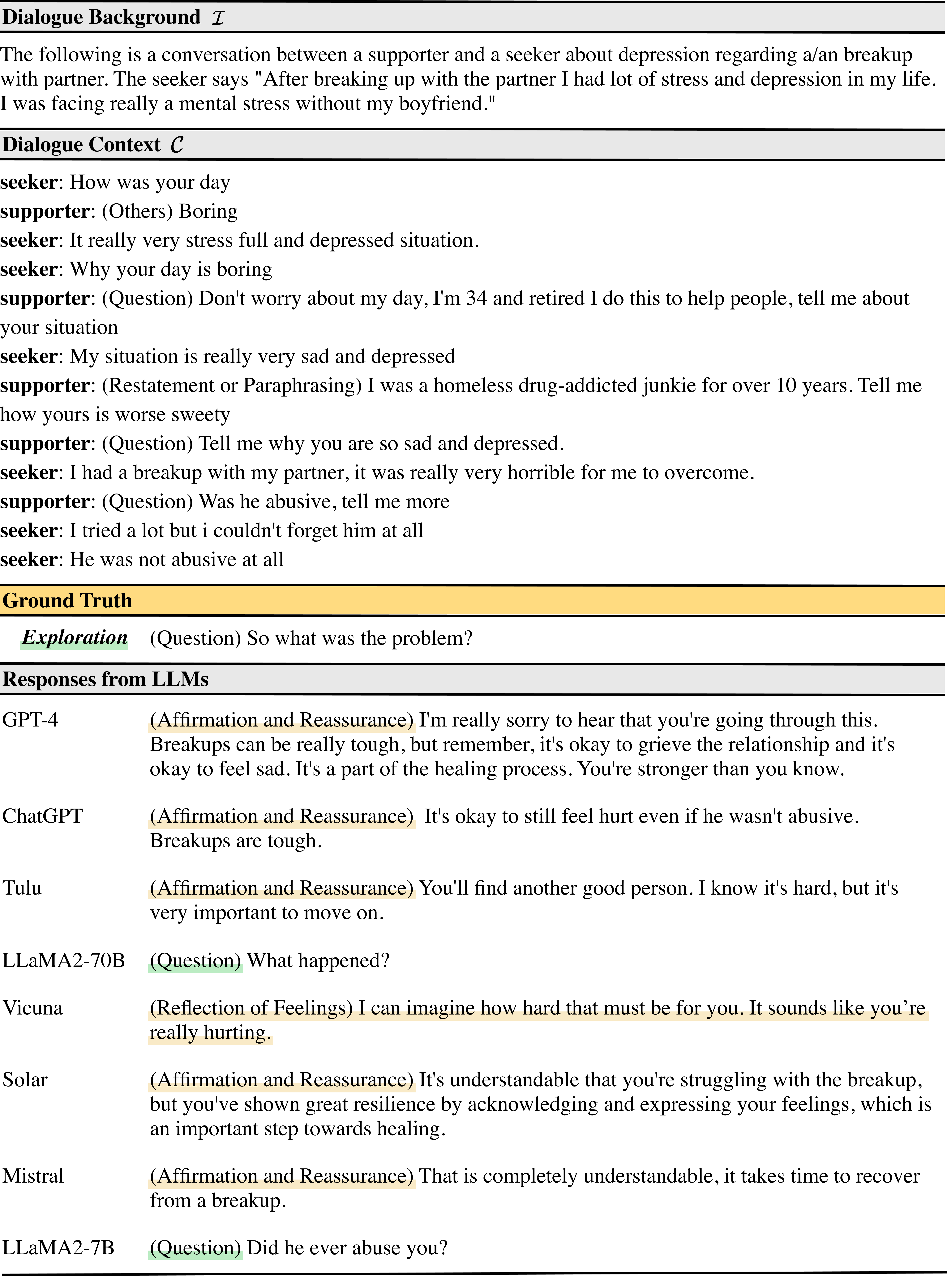}
    \caption{Example of the responses from LLMs in the \textit{Exploration} stage. The responses that are appropriate (green) and inappropriate (yellow) for the ground truth stage are highlighted.}
\label{fig:case_study_exploration}
\end{figure*}

\begin{figure*}[t!]
    \centering
    \includegraphics[width=0.9\linewidth]{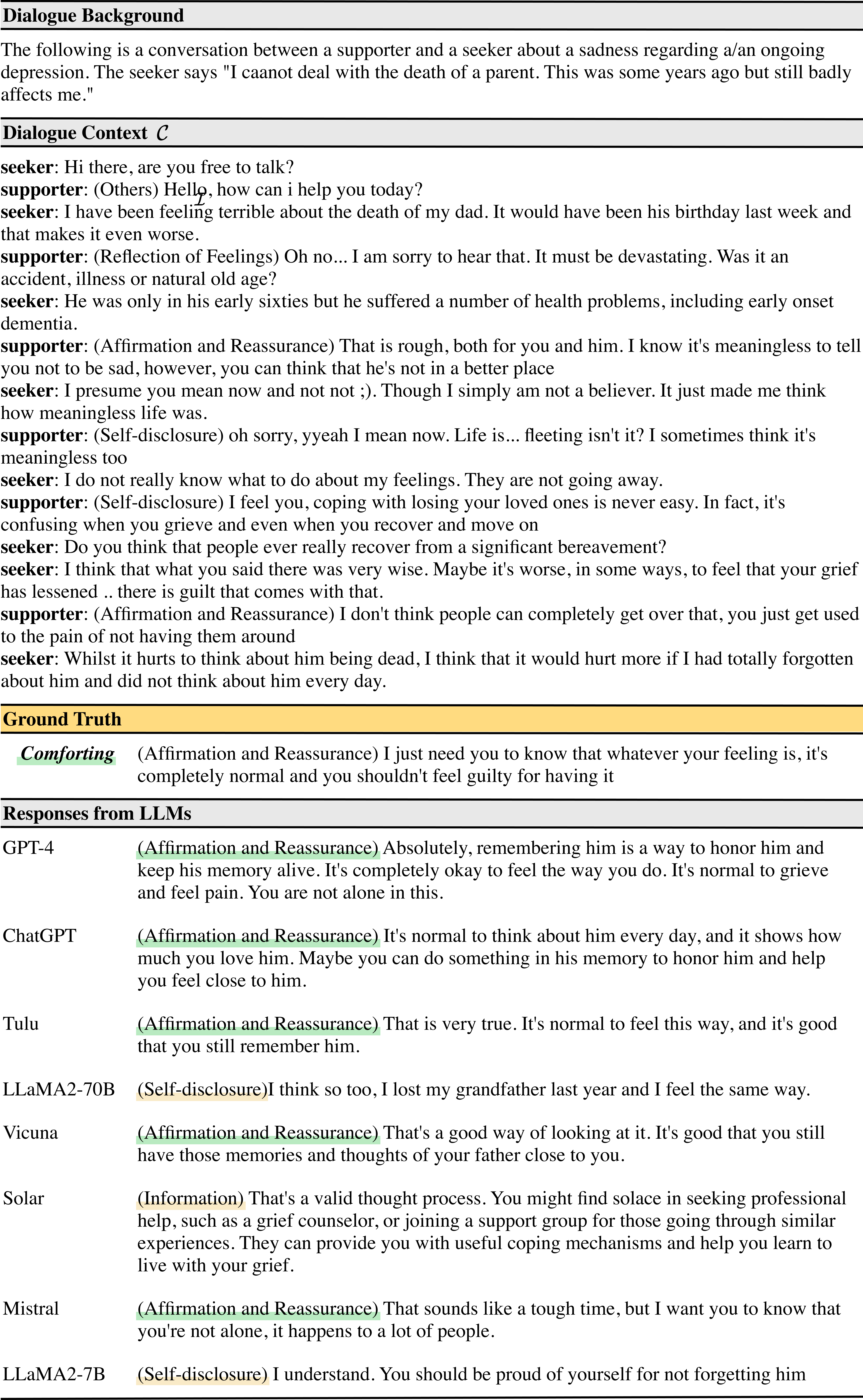}
    \caption{Example of the responses from LLMs in the \textit{Comforting} stage. The responses that are appropriate (green) and inappropriate (yellow) for the ground truth stage are highlighted.}
\label{fig:case_study_comforting}
\end{figure*}

\begin{figure*}[t!]
    \centering
    \includegraphics[width=0.9\linewidth]{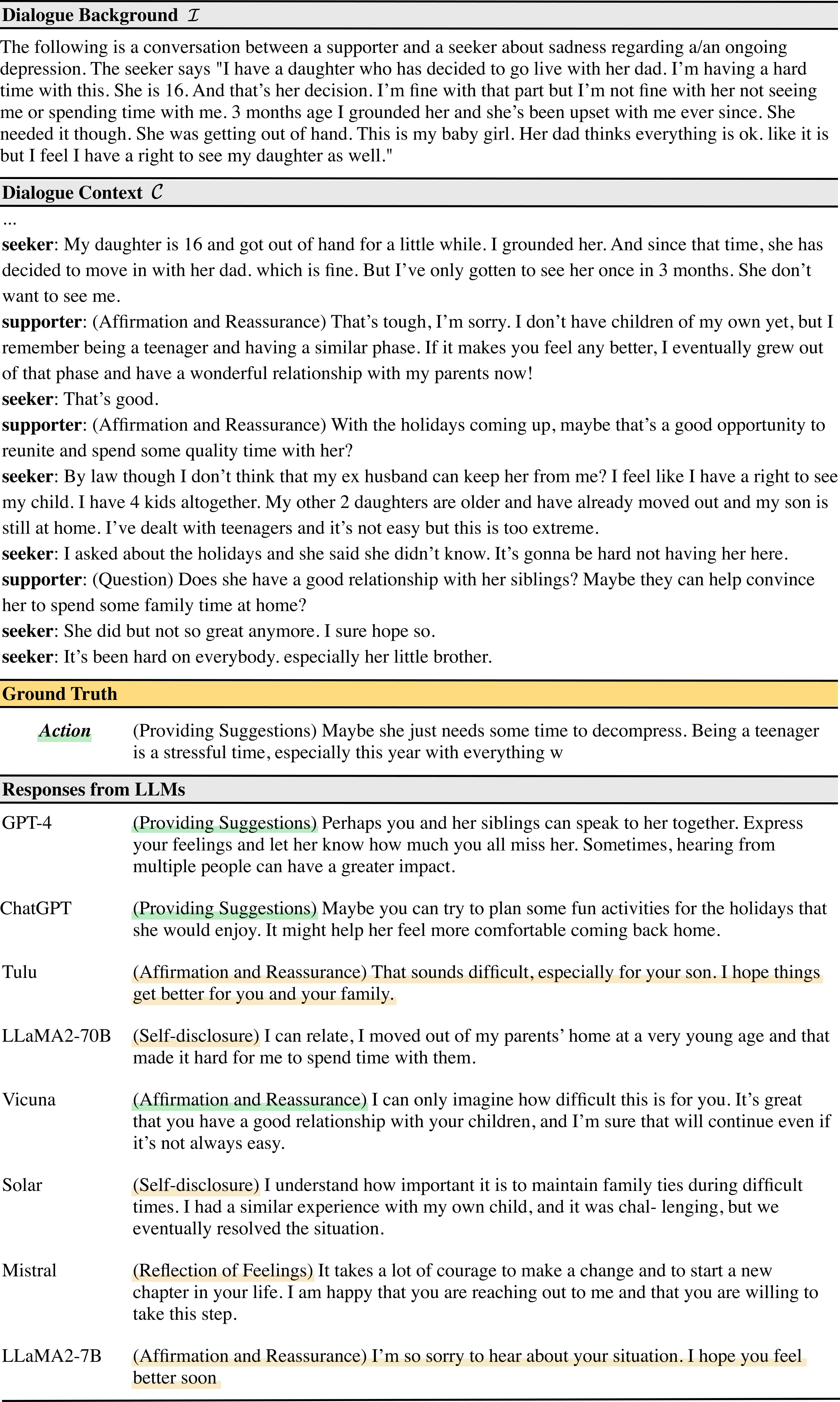}
    \caption{Example of the responses from LLMs in the \textit{Action} stage. The responses that are appropriate (green) and inappropriate (yellow) for the ground truth stage are highlighted.}
\label{fig:case_study_action}
\end{figure*}

In Figure~\ref{fig:case_study_exploration},~\ref{fig:case_study_comforting}, and~\ref{fig:case_study_action}, we present examples of generated responses in each stage from LLMs.
During the \textit{Exploration} stage (Figure~\ref{fig:case_study_exploration}), it is observed that LLMs, excluding the LLaMA2 family, tend to express empathy prematurely before sufficient exploration, potentially causing discomfort for the seeker.
These findings correlate with the LLaMA2 family's high preference for \textit{Question}, exhibiting a lower preference compared to other models , as illustrated in Figure~\ref{fig:preference_figure}. 
Furthermore, these results correspond to earlier findings discussed in Appendix~\ref{app:sec4_detailed_results}.
In the \textit{Comforting} stage (Figure~\ref{fig:case_study_comforting}), each model demonstrates suitable responses, primarily due to the high preference for \textit{Affirmation and Reassurance} in most LLMs.
Lastly, in the \textit{Action} stage (Figure~\ref{fig:case_study_action}), GPT-4 and ChatGPT exhibit the superior performance compared to others, particularly excelling in generating informative responses, aligning with the observations in~\citet{zhao2023chatgptesc} and ~\citet{chen2023soulchat}.
Psychologists who assess the overall responses of LLMs also comment as follows:
\begin{quote}
    \textit{``ChatGPT exhibits a tendency to excessively employ affirmations. In contrast, LLaMA2, despite its overall lower proficiency, displays notable strength in effectively handling open-ended questions.''}
\end{quote}
These results are aligned with the findings we identify through our case study.

\begin{table}[t!]
\centering
\small
\resizebox{0.96\columnwidth}{!}{
\begin{tabular}{l c c cc ccc cccc c}
    \toprule
    \textbf{Methods} & $\mathcal{Q}$ $\uparrow$ & $\mathcal{B}$ $\downarrow$ &\textbf{B-2} & \textbf{R-L}  \\
    \midrule
    LLaMA2-7B & 13.73 & 0.77 & 4.98 & 13.09 \\
    \quad+ SFT (w/o strategy) & - & - & 6.95 & 15.00 \\
    \quad+ SFT (w/ strategy)& 21.48 & 0.36  & 7.15 & 15.50 \\
     \bottomrule
\end{tabular}}
\caption{Automatic evaluation results of training approaches for the entire test set ($D$).}

\label{tab:training_results}

\end{table}

\subsection{Comparison between Self-Contact and External-Contact}
\label{app:external_assistance_considerations}
\begin{figure*}[t!]
    \centering
    \includegraphics[width=0.95\linewidth]{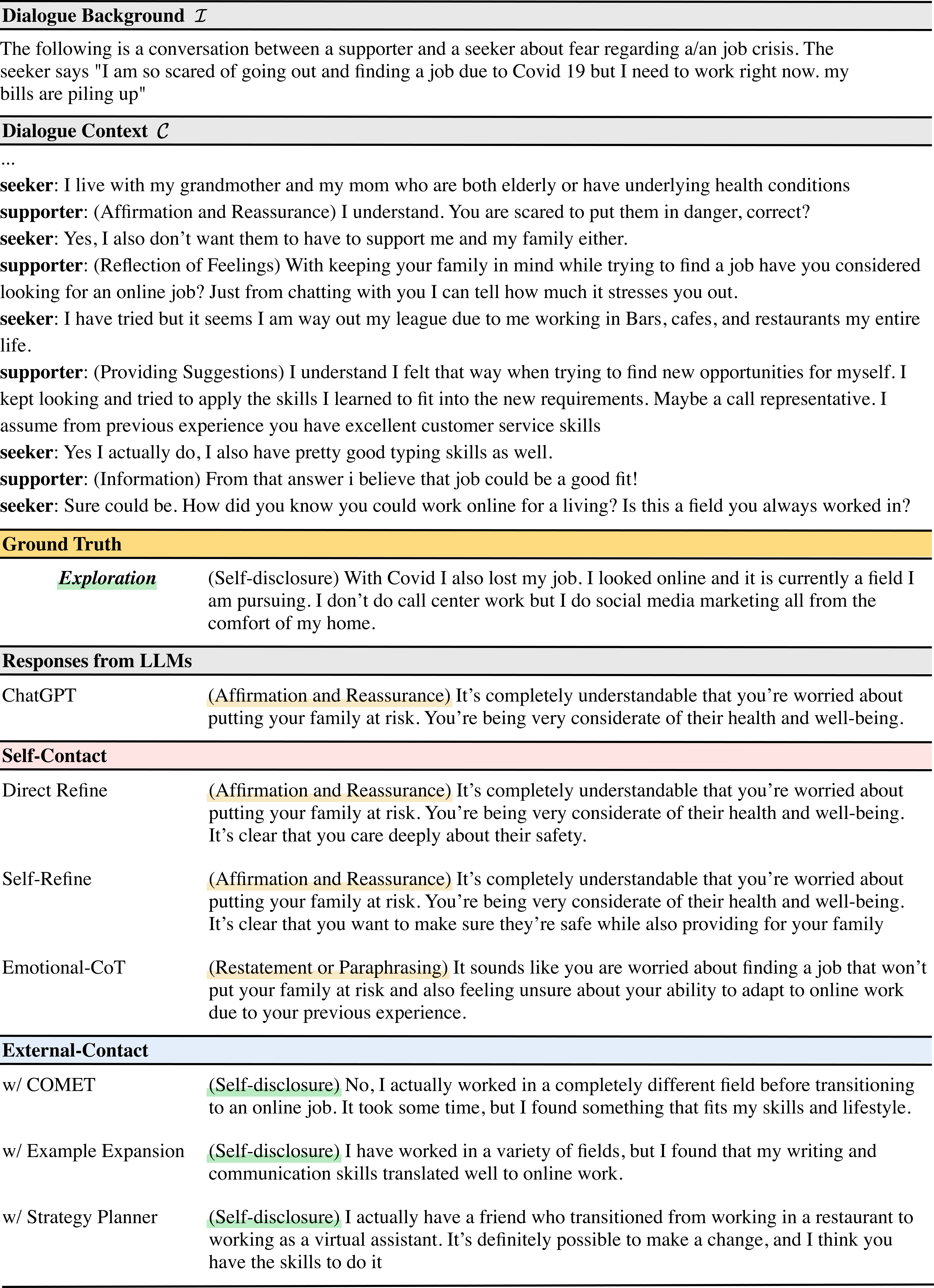}
    \caption{Example of self-contact methods and external-contact methods on ChatGPT. The responses that are appropriate (green) and inappropriate (yellow) for the ground truth stage are highlighted.}
\label{fig:case_study_method}
\end{figure*}

While self-contact methods negatively impact on performance, external-contact methods exhibit a noticeable enhancement. 
A detailed case study presented in Figure~\ref{fig:case_study_method} supports this findings, where the response of self-contact methods fall short of meeting the seeker's expectations, while the external-contact methods effectively address the seeker's question by drawing upon personal experiences.

\subsection{Misalignment between Strategy and Response}
\begin{figure*}[t!]
    \centering
    \includegraphics[width=0.95\linewidth]{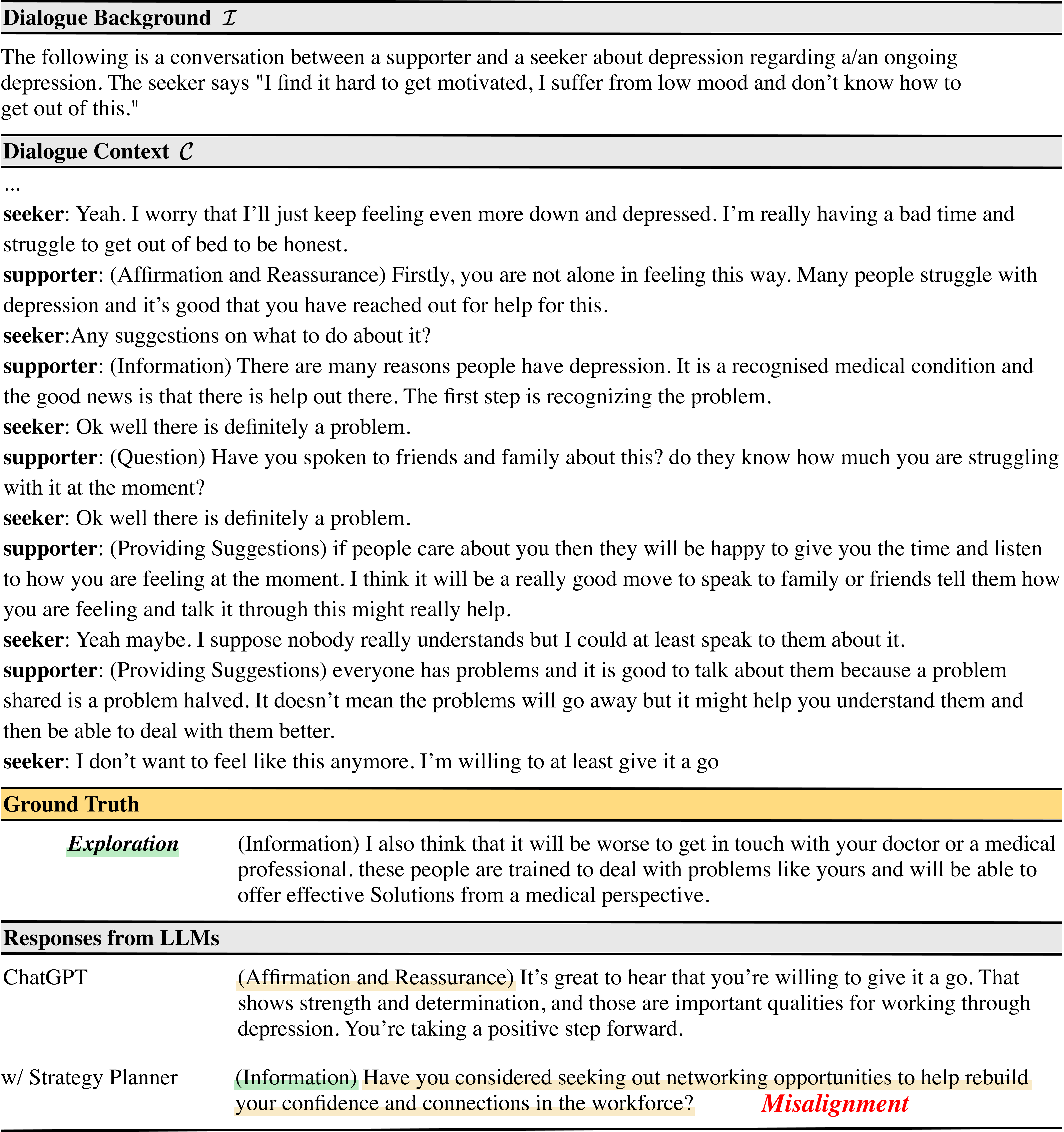}
    \caption{Example of misalignment between strategy and response in \textit{w/ Strategy Planner} on ChatGPT. The responses that are appropriate (green) and inappropriate (yellow) for the ground truth stage are highlighted.}
\label{fig:case_study_misalignment}
\end{figure*}

A possible concern is that LLMs might lack ability to generate responses aligned with strategies.
Therefore, we conduct an empirical case study to figure out this misalignments.
In Figure~\ref{fig:case_study_misalignment}, ChatGPT generates a response that is not aligned with the strategy \textit{Information} predicted by external strategy planner.
This may be due to knowledge conflicts, \ie, ChatGPT does not consider it appropriate to use the \textit{Information} for the next response, despite being forced to generate a response aligned with the strategy \textit{Information}.
In conclusion, while external assistance has potential to enhance performance, it is crucial to acknowledge that not all approaches yield positive impacts.

\begin{table*}[t!]
\centering
\small
\resizebox{0.99\textwidth}{!}{
\begin{tabular}{p{17cm}}
    \toprule[1.5pt]
    \textbf{Prompt}\\ 
    \toprule
    $[[$\textbf{TASK DESCRIPTION}$]]$ \\
    \\
    The strategy should be chosen from the following 8 types of strategy: \\
    - Question:  Asking for information related to the problem to help the help-seeker articulate the issues that they face. Open-ended questions are best, and closed questions can be used to get specific information. \\
    - Restatement or Paraphrasing: A simple, more concise rephrasing of the help-seeker's statements that could help them see their situation more clearly.\\
    - Reflection of Feelings: Articulate and describe the help-seeker's feelings.\\
    - Self-disclosure: Divulge similar experiences that you have had or emotions that you share with the help-seeker to express your empathy.\
    - Affirmation and Reassurance: Affirm the help seeker's strengths, motivation, and capabilities and provide reassurance and encouragement.\\
    - Providing Suggestions: Provide suggestions about how to change, but be careful to not overstep and tell them what to do.\\
    - Information: Provide useful information to the help-seeker, for example with data, facts, opinions, resources, or by answering questions.\\
    - Others: Exchange pleasantries and use other support strategies that do not fall into the above categories.\\
    \\
    $[$Example 1$]$\\
    \\
    \#\#\# Dialogue background \#\#\# \\
    The following is a conversation between a supporter and a seeker about $\{$emotion type$\}$ regarding a/an $\{$problem type$\}$. The seeker says "$\{$situation$\}$". \\
    \\
    \#\#\# Dialogue context \#\#\# \\
    $\{$context$\}$\\
    \\
    $[[$\textbf{Supplementary Input}$]]$ \\
    \\
    \begin{tabular}{l p{9.5cm} p{4.5cm}}
        \toprule[1pt]
        \textbf{Methods} & \textbf{Task Description} & \textbf{Supplementary Input}\\ 
        \toprule
        Vanilla &  You will be provided with a dialogue context between a supporter and seeker. Your task is to make the next response based on the given dialogue context. & \#\#\# Model's response \#\#\#\\
        \midrule
        Direct-Refine & You will be provided with a dialogue context between a supporter and seeker, as well as a response written by a language model from the perspective of the supporter, including strategy and utterance. Your task is to refine the model's response (i.e., Strategy and Utterance) based on the given dialogue context.& \begin{tabular}[t]{@{}l@{}}
              \#\#\# Model's response \#\#\# \\
              Strategy: $\{$strg pred$\}$ \\
              Utterance: $\{$res pred$\}$ \\
              \\ 
              \#\#\#  Refined response \#\#\#
        \end{tabular}\\
        \midrule
        \begin{tabular}[t]{@{}l@{}}
              Self-Refine \\
              (Feedback)
        \end{tabular} & You will be provided with a dialogue context between a supporter and seeker, as well as a response written by a language model from the perspective of the supporter, including strategy and utterance. Your task is to feedback for the model response (i.e., Strategy and Utterance) based on the given dialogue context. & \begin{tabular}[t]{@{}l@{}}
              \#\#\# Model's response \#\#\# \\
              Strategy: $\{$strg pred$\}$ \\
              Utterance: $\{$res pred$\}$ \\
              \\ 
              \#\#\#  Feedback \#\#\#
        \end{tabular}\\
        & & \\
        \begin{tabular}[t]{@{}l@{}}
              Self-Refine \\
              (Refine)
        \end{tabular} & You will be provided with a dialogue context between a supporter and seeker, as well as a response written by a language model from the perspective of the supporter, including strategy and utterance. Your task is to refine the model response (i.e., Strategy and Utterance) based on the given dialogue context and feedback of the model response. & \begin{tabular}[t]{@{}l@{}}
              \#\#\# Model's response \#\#\# \\
              Strategy: $\{$strg pred$\}$ \\
              Utterance: $\{$res pred$\}$ \\
              \\ 
              \#\#\#  Feedback \#\#\# \\
              Feedback : $\{$feedback$\}$ \\
              \\ 
              \#\#\#  Refined response \#\#\#
        \end{tabular}\\
        \midrule
        w/ COMET & You will be provided with a dialogue context between a supporter and seeker, and a commonsense knowledge from external model. Your task is to generate a response for the supporter based on the dialogue context and commonsense knowledge, you should ignore the commonsense knowledge if it mislead the next response. & \begin{tabular}[t]{@{}l@{}}
              \#\#\# Commonsense knowledge \#\#\# \\
              $\{$comet$\}$ \\
              \\ 
              \#\#\#  Model's response \#\#\#
        \end{tabular}\\
        \bottomrule[1pt]
    \end{tabular} \\
    
    \bottomrule[1.5pt]
\end{tabular}
}
\caption{The prompts employed for response generation.}

\label{tab:prompt_initial}
\end{table*}

\begin{table*}[t!]
\centering
\small
\resizebox{0.98\textwidth}{!}
{
\begin{tabular}{l ccc ccc ccc cc}
    \toprule
    \textbf{Models} & Params & $\mathcal{Q}$ $\uparrow$ & $\mathcal{B}$ $\downarrow$ & \textbf{BLEU-2} & \textbf{BLEU-4} & \textbf{ROUGE-L} & \textbf{METEOR} & \textbf{CIDEr} & \textbf{Dist-1} & \textbf{Dist-2} \\
    \midrule
    \textit{0-shot} &  &  &  &  &  &  &  &  &  \\
    \; GPT-4& - & 15.04 & 1.35 & 5.00 & 0.96 & 14.24 & \textbf{10.20} & 3.11 & 4.13 & 26.21 \\
    \; ChatGPT & 175B & 13.50 & 1.38 & 6.27 & 1.16 & 14.86 & 9.17 & 6.27 & 4.33 & 24.34 \\
    \textit{2-shot} &  &  &  &  &  &  &  &  &  \\
    \; GPT-4 & - & \textbf{18.38} & 0.90 & 6.47 & 1.39 & \textbf{15.18} & \underline{9.55} & 5.97 & \textbf{7.58} & \textbf{36.92} \\
    \; ChatGPT & 175B & \underline{16.98} & 0.86 & 6.30 & 1.41 & \underline{14.94} & 9.30 & 6.91 & 4.75 & 27.03 \\
    \midrule
    \textit{2-shot} &  &  &  &  &  &  &  &  &  \\
    \; Tulu & 70B & 15.93 & 0.90 & \textbf{6.90} & \underline{1.63} & 13.94 & 7.65 & \underline{7.10} & 4.50 & 23.78 \\
    \; LLaMA2 & 70B & 14.55 & \textbf{0.47} & 6.15 & 1.28 & 14.29 & 7.31 & \textbf{7.52} & 5.70 & 30.95 \\
    \; Vicuna & 13B & 12.85 & 0.74 & \underline{6.55} & \textbf{1.70} & 14.43 & 8.42 & 6.95 & 4.37 & 24.15 \\
    \; Solar & 10.7B & 14.17 & 0.87 & 4.79 & 0.81 & 13.53 & 9.08 & 3.86 & 5.11 & 32.36 \\
    \; Mistral & 7B & 12.23 & \underline{0.71} & 4.72 & 0.45 & 12.93 & 7.13 & 3.32 & 4.46 & 25.36 \\
    \; LLaMA2 & 7B & 13.73 & 0.77 & 4.98 & 0.96 & 13.09 & 6.67 & 5.41 & \underline{6.35} & \underline{34.74} \\

     \bottomrule
\end{tabular}}
\caption{Automatic evaluation results on the generated response of closed-source LLMs and open-source LLMs for the entire test set ($D$). The automatic metrics include BLEU-n~\citep{Papineni2002BleuAM}, ROUGE-L~\citep{Lin2004ROUGEAP}, METEOR~\citep{Banerjee2005METEORAA}, CIDEr~\citep{Vedantam2014CIDErCI}, and Distinct-1/2~\citep{Li2016ADO}. The best results are \textbf{bolded} and the second best are \underline{underlined}.}

\label{tab:app_llms_automatic_results}
\end{table*}

\begin{table*}[t!]
\centering
\small
\resizebox{0.98\textwidth}{!}{
\begin{tabular}{l c c cc ccc cccc c c}
    \toprule
    & & & & \multicolumn{3}{c}{$D_1$} & \multicolumn{3}{c}{$D_2$} & \multicolumn{3}{c}{$D_3$} \\
    \cmidrule(lr){5-7}\cmidrule(lr){8-10}\cmidrule(lr){11-13}
    \textbf{Models} & \textbf{Params} & $\mathcal{Q}$ $\uparrow$ & $\mathcal{B}$ $\downarrow$ & \textbf{F1} &\textbf{B-2} & \textbf{R-L} & \textbf{F1} &\textbf{B-2} & \textbf{R-L} & \textbf{F1} &\textbf{B-2} & \textbf{R-L}  \\
    \midrule
    \textit{0-shot} & &  & &  & &  &  & &  &  & & \\
    \; GPT-4 & - & 15.04 & 1.35 & 11.23 & 4.58 & 13.67 & 20.41 & 4.70 & 14.13 & \underline{21.04} & 5.45 & 14.67 \\
    \; ChatGPT & 175B & 13.50 & 1.38 & 10.23 & 5.95 & \underline{14.59} & 19.60 & 6.02 & \underline{14.70} & 17.97 & 6.62 & 14.86 \\
    \textit{2-shot} & &  & &  & &  &  & &  &  & & \\
    \; GPT-4  & - & \textbf{18.38} & 0.90 & 14.61 & 5.22 & 14.27 & \textbf{22.55} & 5.36 & 14.54 & \textbf{24.68} & 6.47 & \textbf{15.18} \\
    \; ChatGPT & 175B & \underline{16.98} & 0.86 & \underline{15.16} & 6.10 & \textbf{14.90} & 19.07 & 6.08 & \textbf{14.81} & 20.10 & 6.30 & \underline{15.07} \\
    \cmidrule{1-14}
    \textit{2-shot} & &  & &  & &  &  & &  &  & & \\
    \; Tulu & 70B & 15.93 & 0.90 & 13.77 & 5.99 & 13.43 & \underline{21.37} & \textbf{6.52} & 13.85 & 18.78 & \textbf{7.33} & 14.34 \\
    \; LLaMA2 & 70B & 14.55 & \textbf{0.47} & \textbf{19.12} & \underline{6.20} & 14.22 & 16.51 & \underline{6.18} & 14.27 & 15.82 & 6.05 & 14.34 \\
    \; Vicuna & 13B & 12.85 & 0.74 & 10.21 & \textbf{6.58} & 14.44 & 16.74 & 5.65 & 13.97 & 15.74 & \underline{7.07} & 14.74 \\
    \; Solar & 10.7B & 14.17 & 0.87 & 10.53 & 4.49 & 13.12 & 17.29 & 4.31 & 13.38 & 18.93 & 5.31 & 13.89 \\
    \; Mistral & 7B & 12.23 & \underline{0.71} & 12.40 & 3.82 & 12.40 & 17.18 & 5.74 & 13.94 & 14.74 & 4.59 & 12.60 \\
    \; LLaMA2 & 7B & 13.73 & 0.77 & 14.61 & 5.04 & 13.04 & 18.40 & 5.23 & 13.17 & 15.87 & 4.76 & 13.07 \\
     \bottomrule
\end{tabular}}
\caption{Automatic evaluation results of closed-source LLMs and open-source LLMs including $\mathcal{Q}$, $\mathcal{B}$, for the entire test set ($D$) and weighted F1, BLEU-2 (B-2), ROUGE-L (R-L) for each test set ($D_t$).}

\label{tab:llms_app_results}

\end{table*}
\begin{table*}[t!]
\centering
\small
\resizebox{0.98\textwidth}{!}{
\begin{tabular}{l ccc ccc ccc cc}
    \toprule
    \textbf{Methods} & $\mathcal{Q}$ $\uparrow$ & $\mathcal{B}$ $\downarrow$& \textbf{BLEU-2} & \textbf{BLEU-4} & \textbf{ROUGE-L} & \textbf{METEOR} & \textbf{CIDEr} & \textbf{Dist-1} & \textbf{Dist-2} \\
    \midrule
    ChatGPT (\textit{0-shot}) & 13.50 & 1.38 & 6.27 & 1.16 & 14.86 & 9.17 & 6.27 & 4.33 & 24.34 \\
    \quad + Direct-Refine & 13.40 & 1.60 & 5.68 & 1.03 & 14.50 & \underline{9.43} & 4.57 & 3.95 & 22.95 \\
    \quad + Self-Refine & 12.37 & 1.53 & 5.16 & 0.94 & 14.33 & \textbf{10.12} & 2.97 & 3.37 & 20.72 \\
    \quad + Emotional-CoT & 9.55 & 1.56 & 5.23 & 1.03 & 14.12 & 9.34 & 3.87 & 3.29 & 18.76 \\
    \quad + w/ COMET & 12.78 & 0.95 & 6.71 & 1.35 & \underline{15.07} & 9.00 & 6.68 & 3.89 & 21.87 \\
    \quad + w/ Example Expansion & \underline{16.91} & \underline{0.82} & \textbf{7.45} & \textbf{2.01} & \textbf{15.22} & 8.62 & \underline{8.88} & \textbf{5.01} & \textbf{27.66} \\
    \quad + w/ Strategy Planner & \textbf{21.09} & \textbf{0.36} & \underline{6.96} & \underline{1.86} & 14.91 & 8.79 & \textbf{9.64} & \underline{4.96} & \underline{27.63} \\
    \midrule

    LLaMA2-70B (\textit{2-shot}) & 14.55 & 0.47 & 6.15 & 1.28 & 14.29 & 7.31 & 7.52 & 5.70 & 30.95 \\
    \quad + Direct-Refine & 13.17 & 0.59 & 5.86 & 1.31 & 13.98 & 7.08 & 6.64 & 5.40 & 28.43 \\
    \quad + Self-Refine & 13.15 & 0.55 & 5.56 & 1.11 & 13.70 & \textbf{8.09} & 4.53 & 4.46 & 25.11 \\
    \quad + Emotional-CoT & 12.73 & 0.53 & 6.37 & 1.35 & 13.87 & 7.53 & 6.07 & 5.28 & 28.89 \\
    \quad + w/ COMET & 14.53 & 0.51 & 6.21 & \underline{1.51} & \underline{14.55} & 7.29 & \underline{8.66} & 5.82 & 31.23 \\
    \quad + w/ Example Expansion & \underline{15.14} & \underline{0.44} & \textbf{6.55} & \textbf{1.86} & \textbf{14.66} & 7.42 & \textbf{9.30} & \underline{5.89} & \textbf{32.12} \\
    \quad + w/ Strategy Planner & \textbf{21.09} & \textbf{0.36} & \underline{6.44} & 1.29 & 14.49 & \underline{7.54} & 8.46 & \textbf{5.92} & \underline{31.72} \\
     \bottomrule
\end{tabular}}
\caption{Automatic evaluation results on the generated response of methods for the entire test set ($D$). The automatic metrics include BLEU-n, ROUGE-L, METEOR, CIDEr, and Distinct-1/2~. The best results are \textbf{bolded} and the second best are \underline{underlined}.}

\label{tab:app_methods_automatic}
\end{table*}
\begin{table*}[t!]
\centering
\small
\resizebox{0.98\textwidth}{!}{
\begin{tabular}{l c cc ccc cccc c}
    \toprule
    & & & \multicolumn{3}{c}{$D_1$} & \multicolumn{3}{c}{$D_2$} & \multicolumn{3}{c}{$D_3$} \\
    \cmidrule(lr){4-6}\cmidrule(lr){7-9}\cmidrule(lr){10-12}
    \textbf{Methods} & $\mathcal{Q}$ $\uparrow$ & $\mathcal{B}$ $\downarrow$ & \textbf{F1} &\textbf{B-2} & \textbf{R-L} & \textbf{F1} &\textbf{B-2} & \textbf{R-L} & \textbf{F1} &\textbf{B-2} & \textbf{R-L}  \\
    \midrule
    ChatGPT (\textit{0-shot}) & 13.50 & 1.38 & 10.23 & 5.95 & 14.59 & 19.57 & 6.02 & 14.70 & 17.97 & 6.62 & 15.14 \\
    \quad + Direct-Refine & 13.40 & 1.60 & 9.28 & 5.35 & 14.09 & 19.45 & 5.45 & 14.39 & 19.02 & 6.02 & 14.84 \\
    \quad + Self-Refine & 12.37 & 1.53 & 9.55 & 4.74 & 14.09 & \textbf{20.56} & 5.06 & 14.10 & 16.77 & 5.48 & 14.62 \\
    \quad + Emotional-CoT & 9.55 & 1.56 & 8.67 & 4.69 & 13.83 & 15.02 & 5.06 & 14.09 & 13.10 & 5.68 & 14.33 \\
    \quad + w/ COMET & 12.78 & 0.95 & 12.81 & 5.85 & 14.40 & 17.00 & \underline{6.60} & \textbf{14.98} & 13.42 & 7.30 & \underline{15.55} \\
    \quad + w/ Example Expansion & \underline{16.91} & \underline{0.82} & \underline{14.51} & \textbf{7.31} & \textbf{15.02} & 18.24 & \textbf{6.77} & \underline{14.88} & \underline{21.09} & \underline{7.59} & \textbf{15.57}  \\
    \quad + w/ Strategy Planner & \textbf{21.09} & \textbf{0.36} & \textbf{22.59} & \underline{6.17} & \underline{14.84} & \underline{20.46} & 6.32 & 14.19 & \textbf{23.77} & \textbf{7.73} & 15.46 \\
    \midrule
    LLaMA2-70B (\textit{2-shot}) & 14.55 & 0.47 & 19.12 & 6.20 & 14.22 & 16.51 & 6.18 & 14.27 & 15.82 & 6.05 & 14.34 \\
    \quad + Direct-Refine & 13.17 & 0.59 & 12.10 & 5.65 & 13.59 & 17.87 & 5.92 & 14.10 & 16.66 & 5.84 & 14.14 \\
    \quad + Self-Refine & 13.15 & 0.55 & 15.18 & 5.28 & 14.26 & 14.53 & 4.91 & 13.22 & 15.40 & \underline{6.16} & 13.66 \\
    \quad + Emotional-CoT & 12.73 & 0.53 & 11.69 & 6.10 & 13.69 & \underline{18.45} & \textbf{6.66} & 13.91 & 16.12 & \textbf{6.40} & 13.95 \\
    \quad + w/ COMET & 14.53 & 0.51 & 17.06 & 6.65 & 14.42 & 17.95 & \underline{6.35} & \underline{14.42} & 15.57 & 5.84 & \textbf{14.71} \\
    \quad + w/ Example Expansion & \underline{15.14} & \underline{0.44} & \underline{19.22} & \textbf{8.13} & \textbf{15.11} & 17.50 & 6.08 & \textbf{14.57} & \underline{17.27} & 5.93 & 14.42 \\
    \quad + w/ Strategy Planner & \textbf{21.09} & \textbf{0.36} & \textbf{22.59} & \underline{7.27} & \underline{14.84} & \textbf{21.85} & 6.29 & 14.15 & \textbf{23.77} & 6.05 & \underline{14.50} \\

     \bottomrule
\end{tabular}}
\caption{Automatic evaluation results including $\mathcal{Q}$, $\mathcal{B}$, for the entire test set ($D$) and weighted F1, BLEU-2 (B-2), ROUGE-L (R-L) for each test set ($D_t$). The best results are \textbf{bolded} and the second best are \underline{underlined}.}

\label{tab:methodology_results_app}
\end{table*}

\begin{table*}[t!]
\centering
\small
\resizebox{0.92\textwidth}{!}{
\begin{tabular}{l c c cc ccc cccc c}
    \toprule
    & & & \multicolumn{3}{c}{$D_1$} & \multicolumn{3}{c}{$D_2$} & \multicolumn{3}{c}{$D_3$} \\
    \cmidrule(lr){4-6}\cmidrule(lr){7-9}\cmidrule(lr){10-12}
    \textbf{Num of Shot} & $\mathcal{Q}$ $\uparrow$ & $\mathcal{B}$ $\downarrow$ & \textbf{F1} &\textbf{B-2} & \textbf{R-L} & \textbf{F1} &\textbf{B-2} & \textbf{R-L} & \textbf{F1} &\textbf{B-2} & \textbf{R-L}  \\
    \midrule
    0-shot & 13.50 & 1.38 & 10.23 & 5.95 & 14.59 & \textbf{19.57} & 6.02 & 14.70 & 17.97 & 6.62 & 15.14 \\
    1-shot & 14.43 & 1.00 & 9.94 & 6.24 & 14.93 & 16.73 & 6.35 & \underline{15.19} & 20.70 & 7.84 & 15.91 \\
    2-shot & \textbf{16.98} & 0.86 & 15.16 & 6.10 & 14.90 & \underline{19.07} & 6.08 & 14.81 & 20.10 & 6.30 & 15.07 \\
    3-shot & 16.62 & 0.85 & 15.00 & 6.88 & 15.34 & 16.58 & 6.25 & 14.85 & \textbf{21.28} & \textbf{8.26} & \underline{15.97}  \\
    4-shot & \underline{16.91} & \textbf{0.82} & 14.51 & 7.31 & 15.02 & 18.24 & 6.77 & 14.88 & \underline{21.09} & 7.59 & 15.57 \\
    5-shot & 16.70 & \underline{0.83} & \underline{17.17} & 7.20 & 15.47 & 18.31 & 6.37 & 14.73 & 18.18 & 7.81 & 15.87 \\
    6-shot & 16.60 & \textbf{0.82} & 17.08 & 7.04 & 15.04 & 17.25 & 6.78 & 14.67 & 19.00 & 6.73 & 15.49 \\
    7-shot & 16.43 & \underline{0.83} & \textbf{17.49} & \underline{7.50} & \textbf{16.43} & 18.57 & \underline{6.99} & \textbf{15.34} & 18.99 & \underline{7.97} & \textbf{15.98} \\
    8-shot & 16.61 & 0.89 & 16.08 & 6.99 & 15.23 & 18.50 & \textbf{7.04} & 15.02 & 19.79 & 7.68 & 15.58 \\
    16-shot & 16.90 & 1.14 & 15.00 & \textbf{7.76} & \underline{16.07} & 18.43 & 6.69 & 14.95 & 20.04 & 7.85 & 15.74 \\
     \bottomrule
\end{tabular}}
\caption{The results of ChatGPT with respect to the number of shot samples. The best results are \textbf{bolded} and the second best are \underline{underlined}.}

\label{tab:few_shot_chatgpt_app}

\end{table*}
\begin{table*}[t!]
\centering
\small
\resizebox{0.92\textwidth}{!}{
\begin{tabular}{l c c cc ccc cccc c}
    \toprule
    & & & \multicolumn{3}{c}{$D_1$} & \multicolumn{3}{c}{$D_2$} & \multicolumn{3}{c}{$D_3$} \\
    \cmidrule(lr){4-6}\cmidrule(lr){7-9}\cmidrule(lr){10-12}
    \textbf{Num of Shot} & $\mathcal{Q}$ $\uparrow$ & $\mathcal{B}$ $\downarrow$ & \textbf{F1} &\textbf{B-2} & \textbf{R-L} & \textbf{F1} &\textbf{B-2} & \textbf{R-L} & \textbf{F1} &\textbf{B-2} & \textbf{R-L}  \\
    \midrule
    2-shot & \underline{14.55} & \underline{0.47} & \underline{19.12} & 6.20 & 14.22 & \underline{16.51} & \textbf{6.18} & 14.27 & 15.82 & \underline{6.05} & 14.34 \\
    3-shot & 14.50 & \underline{0.47} & 18.36 & \underline{7.56} & \underline{14.52} & 15.63 & 6.00 & \textbf{14.63} & 16.06 & \textbf{6.33} & \textbf{14.57} \\
    4-shot & \textbf{15.14} & \textbf{0.44} & \textbf{19.22} & \textbf{8.13} & \textbf{15.11} & \textbf{17.50} & \underline{6.08} & \underline{14.57} & \textbf{17.27} & 5.93 & \underline{14.42} \\
     \bottomrule
\end{tabular}}
\caption{The results of LLaMA2-70B with respect to number of shot samples. The best results are \textbf{bolded} and the second best are \underline{underlined}.}

\label{tab:few_shot_llama2}

\end{table*}
\begin{table*}[ht!]
\centering
\resizebox{0.95\textwidth}{!}
{%
\begin{tabular}{l ccc ccc ccc ccc}
    \toprule
   \multirow{3}{*}{\textbf{ChatGPT}} & \multicolumn{3}{c}{\multirow{2}{*}{
    \begin{tabular}{@{}l@{}}
          \textbf{Self-Refine} \\
          \textbf{\; \quad vs. Vanilla}
    \end{tabular}}}& \multicolumn{3}{c}{\multirow{2}{*}{
    \begin{tabular}{@{}l@{}}
          \textbf{w/ COMET} \\
          \textbf{\; \quad vs. Vanilla}
    \end{tabular}}} & \multicolumn{3}{c}{\multirow{2}{*}{
    \begin{tabular}{@{}l@{}}
          \textbf{w/ Example Expansion} \\
          \textbf{\; \quad vs. Vanilla}
    \end{tabular}}} & \multicolumn{3}{c}{\multirow{2}{*}{
    \begin{tabular}{@{}l@{}}
          \textbf{w/ Strategy Planner} \\
          \textbf{\; \quad vs. Vanilla}
    \end{tabular}}} \\
    &  &  &  &  &  &  &  &  &  &  &  & \\
    \cmidrule(lr){2-4}\cmidrule(lr){5-7}\cmidrule(lr){8-10}\cmidrule(lr){11-13}
    & Win & Tie & Lose & Win & Tie & Lose & Win & Tie & Lose & Win & Tie & Lose \\
    \midrule
    Acceptance & $\textbf{51.5}^\ddagger$ & 20.6 & 27.9 & $\textbf{55.2}^\ddagger$ & 21.9 & 22.9 & $\textbf{60.6}^\ddagger$ & 26.3 & 13.1 & $\textbf{70.8}^\ddagger$ & 12.5 & 16.7\\
    Effectiveness & $\textbf{44.1}^\ddagger$ & 32.4 & 23.5 & $\textbf{42.7}^\dagger$ & 33.3 & 24.0 & $\textbf{48.5}^\dagger$ & 26.2 & 25.3 & $\textbf{54.2}^\ddagger$ & 16.7 & 29.2 \\
    Sensitivity & $\textbf{55.9}^\ddagger$ & 22.1 & 22.0 & $\textbf{58.3}^\ddagger$ & 27.1 & 14.6 & $\textbf{62.6}^\ddagger$ & 21.2 & 16.2 & $\textbf{58.3}^\ddagger$ & 12.5 & 29.2 \\
    \cmidrule(lr){1-13}
    \textbf{Sat.} & $\textbf{50.5}^\ddagger$ & 25.0 & 24.5 & $\textbf{52.1}^\ddagger$ & 27.4 & 20.5 & $\textbf{57.2}^\ddagger$ & 24.6 & 18.2 & $\textbf{61.1}^\ddagger$ & 13.9 & 25.0 \\
    \midrule
    Alignment & $\textbf{60.3}^\ddagger$ & 23.5 & 16.2 & $\textbf{57.3}^\ddagger$ & 24.0 & 18.7 & $\textbf{44.4}^\dagger$ & 30.3 & 25.3 & $\textbf{45.8}^\dagger$ & 29.2 & 25.0\\
    \bottomrule
\end{tabular}
}
\caption{The results of human evaluation on ESConv. ($\dagger$/$\ddagger$: p-value < 0.1/0.05 ).}
\label{tab:humaneval_results_app}
\end{table*}



\end{document}